\documentclass[journal]{IEEEtran}

\usepackage{amsmath,amssymb,amsfonts}
\usepackage{booktabs}
\usepackage{multirow}
\usepackage{graphicx}
\usepackage{placeins}
\usepackage{float}
\usepackage[nocompress]{cite}
\usepackage{url}

\newcommand{\x}{\mathbf{x}}
\newcommand{\R}{\mathbb{R}}

\title{SAR-TS-BERT: Physics-Guided Self-Supervised Pre-Training With Polarimetric Features for SAR Time-Series Crop Classification}

\author{Jieqiu~Wang\textsuperscript{*},
        Yuhan~Zhang\textsuperscript{*},
        Bo~Zheng,
        and~Shaoming~Zhang%
\thanks{\textsuperscript{*}Jieqiu Wang and Yuhan Zhang contributed equally to this work (co-first authors). \emph{(Corresponding author: Shaoming Zhang.)}}%
\thanks{Jieqiu Wang, Yuhan Zhang, Bo Zheng, and Shaoming Zhang are with the College of Surveying and Geo-Informatics, Tongji University, Shanghai 200092, China (e-mail: 2511816@tongji.edu.cn; 2433826@tongji.edu.cn; 2180086@tongji.edu.cn; zhangshaoming@tongji.edu.cn).}%
\thanks{ORCID: Jieqiu Wang, 0009-0001-5585-0171; Shaoming Zhang, 0000-0001-8040-3894.}%
}

\begin{document}
\bstctlcite{IEEEtranBSTcontrol}

\maketitle

\begin{abstract}
Multitemporal synthetic aperture radar (SAR) imagery provides continuous, cloud-free observations of crop phenology for agricultural monitoring in cloudy and rainy regions, yet high-accuracy classification typically requires extensive labeled samples. Although self-supervised pre-training mitigates label dependency in remote sensing, existing frameworks primarily target optical imagery or generic spatial SAR scenes, underutilizing temporal backscatter dynamics and polarimetric scattering mechanisms. We propose SAR-TS-BERT, a physics-guided BERT-style self-supervised pre-training framework for SAR time-series crop classification. SAR-TS-BERT integrates backscatter intensities and polarimetric ratio features (VV, VH, and VV/VH) with dual-polarimetric Cloude--Pottier decomposition parameters ($H/\alpha/A$) into a six-band input representation through two core designs: 1) a physics-guided span masking strategy (SpanPhysics) that adaptively selects contiguous masked spans based on temporal backscatter change rates to prioritize key phenological transitions; and 2) a dual-branch cross-modal fusion encoder that models backscatter and polarimetric decomposition features through separate Transformer branches followed by bidirectional cross-attention fusion. On independent test sets in Estonia (EuroCropsML) and the USA (California CDL), SAR-TS-BERT achieves overall accuracies (OAs) of 96.78\% and 87.29\%, respectively, and attains a mean monthly OA of 99.37\% on Chongming Island, Shanghai. On the Estonia benchmark, pre-training improves OA and Macro F1 by 1.17 and 1.59 percentage points over training from scratch under full supervision, and by 2.58 and 5.19 percentage points using only 1\% labeled data. Incorporating SAR scattering physics into temporal self-supervised pre-training effectively enhances crop classification accuracy, parcel-level mapping consistency, and few-shot transferability.
\end{abstract}


\section{Introduction}
\label{sec:introduction}

\IEEEPARstart{A}{ccurate} and timely crop-type mapping is fundamental to agricultural monitoring, planting structure inventory, yield estimation, and food security assessment~\cite{planque2021,agrifm2025}. With the continuous advancement of Earth observation capabilities, satellite image time series (SITS) have become a primary data modality for large-scale crop identification, as sequential acquisitions capture distinct phenological trajectories across growth stages~\cite{garnot2020pse,russwurm2020selfattention,sitsbert2021}. Among multi-source remote sensing modalities, synthetic aperture radar (SAR) provides all-weather, day-and-night imaging that bridges observation gaps caused by persistent cloud cover in optical time series~\cite{harfenmeister2021,coherence_backscatter2024}. In particular, the Sentinel-1 mission delivers open-access C-band dual-polarimetric SAR imagery with high revisit frequency, establishing a reliable data source for operational multitemporal crop monitoring~\cite{planque2021,winterwheat_pol2024}.

Effective representation learning for SAR time-series crop classification relies on both temporal dynamics and physical scattering properties. While deep learning models extract temporal crop signatures in an end-to-end manner, achieving high accuracy typically demands extensive labeled samples. Because field surveys and visual interpretation are costly and labor-intensive, ground-truth labels remain scarce across many geographical regions and minor crop types~\cite{sitsbert2021,presto2023}. Self-supervised learning (SSL) mitigates this annotation bottleneck by formulating pretext tasks on unlabeled data, demonstrating strong transferability in optical SITS and multimodal remote sensing~\cite{sitsbert2021,satmae2022,croma2023}. Concurrently, multitemporal SAR observations encode rich physical scattering information beyond raw amplitude. Co-polarized (VV) and cross-polarized (VH) channels record backscatter responses across orthogonal polarization states, whereas polarimetric decomposition parameters characterize the underlying scattering mechanisms of vegetation canopies and soil surfaces. In the dual-polarimetric Cloude--Pottier $H/\alpha/A$ decomposition, polarimetric entropy ($H$), mean scattering angle ($\alpha$), and anisotropy ($A$) quantify scattering randomness, the transition between co- and cross-polarized dominant scattering mechanisms, and the relative contrast between secondary and primary scattering components, respectively. These polarimetric descriptors complement backscatter intensity and separate crop types that exhibit similar amplitude trajectories but distinct structural scattering mechanisms~\cite{cloude_pottier1997,harfenmeister2021,qu2020cropmapping,poldecomp_crop2023}.

Existing studies on SAR and SITS crop classification have explored diverse modeling paradigms. Supervised machine learning and deep sequence models typically ingest multitemporal VV, VH, and VV/VH series using Random Forests (RF), recurrent neural networks (RNNs), temporal convolutional networks (TCNs), or self-attention mechanisms to model crop growth dynamics~\cite{planque2021,coherence_backscatter2024,tsuchiya2025}. Several works further incorporate polarimetric covariance matrices or decomposition features to enhance sensitivity to canopy scattering structures~\cite{qu2020cropmapping,poldecomp_crop2023}. In self-supervised pre-training, SITS-BERT introduced BERT-style masked reconstruction to optical satellite time series, SatMAE extended masked autoencoders (MAE) to multispectral and temporal dimensions, and Presto and CROMA investigated multi-sensor and radar--optical representation learning~\cite{sitsbert2021,satmae2022,presto2023,croma2023}. Meanwhile, SAR-oriented foundation models such as SARATR-X, SAR-W-MixMAE, and SARMAE have advanced self-supervised learning for SAR imagery~\cite{saratrx2024,sarwmae2025,sarmae2025}. Nevertheless, three critical limitations remain: 1) most SITS pre-training frameworks are tailored to optical reflectance and treat SAR channels as generic numerical inputs; 2) SAR-specific foundation models primarily focus on spatial scene classification or target recognition rather than phenology-driven multitemporal classification; and 3) polarimetric decomposition parameters are predominantly used via simple feature concatenation in supervised classifiers, rather than being integrated into self-supervised pretext task design, masking strategies, and cross-modal temporal representation learning.

To address these challenges, we propose SAR-TS-BERT, a physics-guided self-supervised pre-training framework for SAR time-series crop classification. Rather than feeding stacked SAR channels into a single generic Transformer, SAR-TS-BERT explicitly decouples backscatter intensity features from polarimetric scattering mechanisms and leverages temporal backscatter dynamics to guide span masking toward informative phenological transition windows. The main contributions of this study are summarized as follows:

\begin{enumerate}
    \item We propose a physics-aware self-supervised representation learning paradigm for dual-polarimetric SAR time series that, for the first time, jointly models $\mathbf{C}_2$-derived Cloude--Pottier polarimetric decomposition parameters ($H/\alpha/A$) and backscatter features (VV, VH, and VV/VH) within a masked temporal pre-training framework, capturing both echo amplitude responses and scattering mechanism transitions from unlabeled sequences.
    \item We design a dual-branch cross-modal fusion encoder that encodes backscatter and polarimetric decomposition trajectories through dedicated Transformer branches and integrates them via bidirectional cross-attention.
    \item We develop a physics-guided contiguous span masking strategy (SpanPhysics) that adaptively samples masked positions based on adjacent-timestep backscatter change rates, prioritizing the reconstruction of rapid phenological transition segments to strengthen temporal dynamic representations.
    \item Comprehensive experiments across three diverse datasets in Estonia, the USA, and Chongming Island (Shanghai) demonstrate that SAR-TS-BERT achieves state-of-the-art performance in SAR time-series crop classification, large-scale spatial mapping, and few-shot transfer learning.
\end{enumerate}

The remainder of this paper is organized as follows. Section~\ref{sec:related_work} reviews related work on SAR polarimetric decomposition, SAR time-series crop classification, and remote sensing self-supervised pre-training. Section~\ref{sec:material} introduces the study areas, datasets, and SAR preprocessing pipelines. Section~\ref{sec:method} details the proposed SAR-TS-BERT framework. Section~\ref{sec:experiments} presents comparative experimental results and ablation studies. Section~\ref{sec:discussion} discusses physical interpretability, cross-dataset mapping reliability, and limitations, and Section~\ref{sec:conclusion} concludes the paper.

\section{Related Work}
\label{sec:related_work}

\subsection{SAR Polarimetric Decomposition and Feature Engineering}

Polarimetric decomposition extracts physically interpretable scattering descriptors from SAR polarimetric covariance or coherency matrices, providing structural information complementary to raw backscatter coefficients. The classic Cloude--Pottier $H/\alpha/A$ decomposition~\cite{cloude_pottier1997} parameterizes the eigen-structure of the polarimetric coherency matrix into three components: polarimetric entropy ($H$), which measures the degree of scattering randomness; mean scattering angle ($\alpha$), which identifies the dominant scattering mechanism among surface, volume, and double-bounce scattering; and anisotropy ($A$), which characterizes the relative importance of secondary scattering mechanisms. Under the reflection symmetry assumption, the Freeman--Durden three-component decomposition~\cite{freeman_durden1998} partitions the covariance matrix into surface, double-bounce, and volume scattering power contributions, while the Yamaguchi four-component decomposition~\cite{yamaguchi4comp2005} incorporates a helix scattering term to accommodate non-reflection-symmetric urban and vegetated environments. While these classical formulations were originally established for quad-polarimetric coherency ($\mathbf{T}_3$) or covariance ($\mathbf{C}_3$) matrices, operational dual-polarimetric missions such as Sentinel-1 Interferometric Wide-swath (IW) mode perform polarimetric decomposition on the $2 \times 2$ covariance matrix $\mathbf{C}_2$ within the VV--VH subspace~\cite{harfenmeister2021,braun2022}.

In agricultural remote sensing, polarimetric decomposition parameters have proven effective for monitoring crop phenology and discriminating crop types. Harfenmeister \emph{et al.}~\cite{harfenmeister2021} demonstrated that Cloude--Pottier parameters derived from dual-polarimetric Sentinel-1 acquisitions track crop growth stages more effectively than using VV and VH intensities alone. Using synthesized quad-polarimetric data, Braun and Offermann~\cite{braun2022} systematically evaluated the polarimetric information content of Sentinel-1 for land cover mapping and confirmed the discriminative value of eigen-decomposition features for vegetation classes. For crop classification, Liu \emph{et al.}~\cite{poldecomp_crop2023} combined multitemporal polarization descriptors with patch-based neural networks, and Zhang \emph{et al.}~\cite{chongming2026} validated the effectiveness of joint backscatter and polarimetric decomposition inputs for crop mapping on Chongming Island. These studies indicate that polarimetric parameters not only expand feature dimensionality but also provide a physical observational perspective on canopy architecture, vegetation water content dynamics, and soil surface states that is distinct from echo amplitude. However, existing approaches predominantly treat polarimetric parameters as handcrafted features for supervised classifiers, leaving their integration into self-supervised pre-training frameworks and end-to-end cross-modal representation learning largely unexplored.

\subsection{SAR Time-Series Crop Classification}

SAR time-series crop classification relies on temporal variations in radar backscattering across phenological stages. Early approaches combined handcrafted temporal summary features with conventional classifiers, such as applying knowledge-based decision rules or Random Forests to Sentinel-1 VV and VH time series for national-scale crop mapping~\cite{planque2021}, or incorporating interferometric coherence and VV/VH ratios to enhance separability across growth phases~\cite{coherence_backscatter2024,winterwheat_pol2024}. Although interpretable, these methods exhibit limited capacity to model complex nonlinear phenological trajectories and regional variations.

With the advancement of deep learning, recurrent networks, temporal convolutions, and attention mechanisms have been widely adopted to model sequential SAR observations directly. Liao \emph{et al.}~\cite{liao2020conv1d} developed a 1D convolutional neural network (Conv1D) for multitemporal RADARSAT-2 polarimetric SAR and VEN$\mu$S data, showing that Conv1D coupled with minimum noise fraction (MNF)-fused polarimetric features outperforms multilayer perceptrons (MLPs), RNNs, and traditional classifiers. Tsuchiya and Sonobe~\cite{tsuchiya2025} evaluated LSTM, bidirectional GRU (Bi-GRU), and attention-enhanced TCN architectures on Sentinel-1 time series, demonstrating that deep sequence models effectively capture long-range temporal dependencies and critical observation windows. Despite their competitive accuracy, supervised deep models remain constrained by their reliance on large-scale labeled training datasets---a persistent bottleneck in operational agricultural monitoring. Recently, AgriFM~\cite{agrifm2025} introduced a multi-source temporal foundation model using MODIS, Landsat-8/9, and Sentinel-2 observations for agricultural mapping. Its optical input configuration differs from the dual-polarimetric SAR time series investigated in this study.

\subsection{Self-Supervised Pre-Training in Remote Sensing}

Self-supervised learning leverages large volumes of unlabeled imagery to learn transferable representations, crystallizing into two primary paradigms in remote sensing: masked autoencoders (MAE) and BERT-style masked sequence reconstruction.

Originally introduced by He \emph{et al.}~\cite{mae2022} for natural images, the MAE paradigm masks a large proportion of input patches and trains an asymmetric encoder--decoder architecture to reconstruct the missing pixels. SatMAE~\cite{satmae2022} extended this framework to multitemporal and multispectral satellite imagery via temporal and spectral group embeddings, achieving strong performance on land cover and crop classification benchmarks. SpectralGPT~\cite{spectralgpt2024} and S2MAE~\cite{s2mae2024} further modeled cross-band and spatial--spectral relationships, while CROMA~\cite{croma2023} learned unimodal and multimodal representations from paired SAR--optical imagery. Our work focuses on masked reconstruction along the temporal dimension to capture sequential phenological dependencies.

Originating in natural language processing~\cite{bert2019}, the BERT paradigm masks selected tokens along a sequence and trains a bidirectional Transformer to predict the corrupted observations from surrounding temporal context. SITS-BERT~\cite{sitsbert2021} pioneered the adaptation of BERT to optical satellite time series by randomly masking observations and reconstructing their spectral values, demonstrating substantial gains in crop classification under limited label budgets. Patnala \emph{et al.}~\cite{bertbimodal2025} extended BERT pre-training to bimodal Sentinel-2 and PlanetScope time series, and Presto~\cite{presto2023} introduced a lightweight pre-trained Transformer supporting multi-sensor time-series inputs. Nevertheless, these frameworks treat SAR backscatter as standard continuous variables without exploiting the physical structure of polarimetric radar signals.

Self-supervised foundation models specifically designed for SAR imagery have also progressed rapidly. SARATR-X, SAR-W-MixMAE, SARMAE, SARCLIP, and large-scale SAR MAE pre-training have advanced SAR representation learning across target recognition, speckle-robust reconstruction, scene classification, and vision--language alignment~\cite{saratrx2024,sarwmae2025,sarmae2025,sarclip2025,largescalemae2023}. However, these spatial foundation models primarily focus on single-date image patches or scene-level semantics, emphasizing spatial texture and target geometry. In contrast, agricultural crop classification requires modeling continuous temporal transitions across phenological stages and capturing how scattering mechanisms evolve over the growing season. Existing SAR pre-training frameworks rarely address these multitemporal phenological dynamics or incorporate polarimetric decomposition features into the pretext task. Motivated by these gaps, we develop SAR-TS-BERT, a physics-guided BERT-style pre-training framework that jointly models backscatter intensity and polarimetric decomposition trajectories to learn physically grounded crop representations.

\section{Materials and Data}
\label{sec:material}

\subsection{Study Areas and Datasets}
\label{sec:datasets}

\begin{figure}[t]
\centering
\includegraphics[width=\columnwidth]{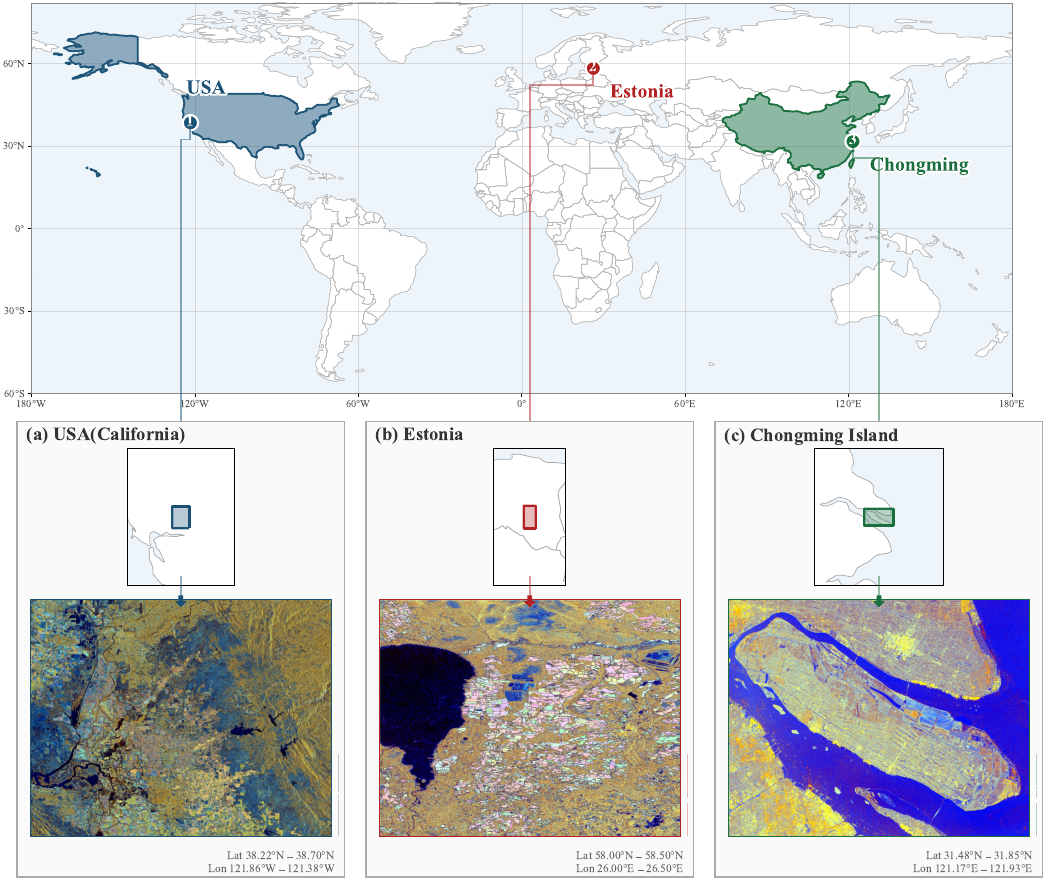}
\caption{Study areas: (a) California, USA (Mediterranean climate); (b) Estonia (Baltic region, cloudy and rainy climate); and (c) Chongming Island, Shanghai (subtropical monsoon climate).}
\label{fig:study_area}
\end{figure}

We evaluate SAR-TS-BERT across three multitemporal SAR datasets spanning distinct geographical regions, climatic zones, and crop compositions, as illustrated in Fig.~\ref{fig:study_area}. All datasets are derived from 10\,m spatial resolution Sentinel-1 acquisitions and uniformly comprise six bands: backscatter features (VV, VH, and VV/VH) and dual-polarimetric Cloude--Pottier decomposition parameters ($H$, $\alpha$, and $A$).

\paragraph{Estonia (EuroCropsML)} Derived from the EuroCropsML benchmark~\cite{eurocropsml2024}, the Estonia dataset covers the national territory of Estonia along the Baltic coast in Northern Europe ($58\text{--}60^\circ$N, $21\text{--}28^\circ$E). Persistent cloud cover and frequent precipitation in this region highlight the operational value of SAR time series for agricultural monitoring. The study area encompasses 5,664 agricultural parcels (corresponding to 540,187 within-parcel pixel-level time-series samples) across nine crop types: clover, legumes harvested green, oats, pasture/meadow/grassland, peas, spring barley, spring wheat, winter wheat, and winter rapeseed. Spanning the entire 2021 growing season with $T=28$ observations at approximately 12-day intervals, the dataset is partitioned at the parcel level into 40\% training (2,266 parcels), 10\% validation (566 parcels), and 50\% testing (2,832 parcels).

\paragraph{USA (Crop Data Layer)} The USA dataset covers the Central Valley of California ($35\text{--}38^\circ$N, $119\text{--}122^\circ$W), a major agricultural production region characterized by a Mediterranean climate with dry, warm summers and mild, wet winters. Crop labels are obtained from the 2024 USDA Cropland Data Layer (CDL) product. The dataset contains approximately 28.95 million (28,950,454) pixel-level samples across eight crop types: corn, winter wheat, alfalfa, tomatoes, grapes, almonds, walnuts, and pistachios. With abundant cloud-free days, this region provides a climatic contrast to Estonia regarding optical data availability. Spanning January to December 2024 with $T=30$ observations, the dataset is partitioned into 20\% training, 5\% validation, and 75\% testing sets.

\paragraph{Chongming Island, Shanghai} Constructed through field surveys, the Chongming Island dataset covers Chongming Island at the mouth of the Yangtze River ($31^\circ$N, $121^\circ$E), which experiences a subtropical monsoon climate with heavy cloud cover during the growing season. The dataset comprises 543 agricultural parcels and 581 field survey points (totaling 451,920 valid monthly pixel-level samples) annotated across 11 crop and land-cover types: rice, wheat, corn, soybean, vegetables, rapeseed, woodland, citrus, rice stubble, plowed land, and weeds. A distinctive feature of this dataset is its monthly annotation protocol: each parcel is labeled monthly from December 2023 to October 2024 (10 valid annotated months excluding February 2024, with $T=25$ Sentinel-1 acquisitions), explicitly capturing crop rotation patterns. The dataset encompasses 13 crop rotation patterns, dominated by wheat--rice (44.28\%) and single-season rice (34.95\%), which together account for 79.23\%, followed by rapeseed--rice (4.87\%), wheat--soybean (4.30\%), and woodland (3.93\%), while the remaining eight patterns account for 7.67\%. This highly imbalanced rotation structure combined with fine parcel granularity constitutes a challenging fine-grained classification scenario.

Table~\ref{tab:dataset_overview} summarizes the primary characteristics of the three datasets, and Tables~\ref{tab:estonia_classes}--\ref{tab:chongming_classes} detail the pixel-level class distributions for each dataset.

\begin{table}[t]
\centering
\caption{Overview of the Three SAR Time-Series Datasets}
\label{tab:dataset_overview}
\begin{tabular}{lccc}
\toprule
Attribute & Estonia & USA & Chongming \\
\midrule
Climate & Cloudy/rainy & Mediterranean & Subtropical monsoon \\
Spatial extent & National & Central Valley & Island \\
Granularity & Parcel-level & Pixel-level & Parcel-level \\
Total samples & 5,664 parcels & $\sim$28.9M pixels & 543 parcels \\
Crop classes & 9 & 8 & 11 \\
Time span & 2021 & 2024 & 2023.12--2024.10 \\
Timesteps ($T$) & 28 & 30 & 25 \\
Revisit interval & $\sim$12 days & $\sim$12 days & $\sim$12 days \\
Split ratio & 40/10/50 & 20/5/75 & 40/10/50 \\
\bottomrule
\end{tabular}
\end{table}

\begin{table}[t]
\centering
\caption{Class Distribution of the Estonia Dataset (Within-Parcel Pixel Samples)}
\label{tab:estonia_classes}
\begin{tabular}{lrr}
\toprule
Crop Type & Samples & Proportion (\%) \\
\midrule
Pasture/meadow/grassland & 260,846 & 48.29 \\
Winter wheat & 70,395 & 13.03 \\
Spring barley & 45,454 & 8.41 \\
Legumes harvested green & 31,266 & 5.79 \\
Clover & 30,765 & 5.70 \\
Peas & 30,389 & 5.63 \\
Oats & 25,312 & 4.69 \\
Winter rapeseed & 24,062 & 4.45 \\
Spring wheat & 21,698 & 4.02 \\
\bottomrule
\end{tabular}
\end{table}

\begin{table}[t]
\centering
\caption{Class Distribution of the USA Dataset}
\label{tab:usa_classes}
\begin{tabular}{lrr}
\toprule
Crop Type & Samples & Proportion (\%) \\
\midrule
Grapes & 9,519,034 & 32.88 \\
Almonds & 5,985,871 & 20.68 \\
Walnuts & 3,599,395 & 12.43 \\
Corn & 3,133,946 & 10.83 \\
Alfalfa & 2,191,909 & 7.57 \\
Tomatoes & 2,187,955 & 7.56 \\
Winter wheat & 2,121,415 & 7.33 \\
Pistachios & 210,929 & 0.73 \\
\bottomrule
\end{tabular}
\end{table}

\begin{table}[t]
\centering
\caption{Crop and Land-Cover Class Distribution of the Chongming Island Dataset (Monthly Pixel Samples)}
\label{tab:chongming_classes}
\begin{tabular}{lrr}
\toprule
Category & Samples & Proportion (\%) \\
\midrule
Rice & 181,435 & 40.15 \\
Wheat & 105,514 & 23.35 \\
Plowed land & 69,486 & 15.38 \\
Rice stubble & 31,972 & 7.07 \\
Woodland & 17,880 & 3.96 \\
Rapeseed & 11,919 & 2.64 \\
Soybean & 11,404 & 2.52 \\
Corn & 7,298 & 1.61 \\
Weeds & 5,592 & 1.24 \\
Citrus & 4,800 & 1.06 \\
Vegetables & 4,620 & 1.02 \\
\bottomrule
\end{tabular}
\end{table}

\subsection{SAR Data Preprocessing}
\label{sec:sar_preprocessing}

Raw Sentinel-1 SAR imagery undergoes systematic preprocessing prior to feature extraction and model training. The preprocessing workflow consists of two parallel pipelines—backscatter coefficient extraction and dual-polarimetric $H/\alpha/A$ decomposition—both executed in ESA SNAP 11.0.0.

\paragraph{Backscatter Coefficient Extraction} Using Ground Range Detected High-Resolution (GRDH) products as input, the pipeline sequentially performs: 1) \emph{precise orbit application}, replacing coarse metadata orbit state vectors with ESA precise orbit ephemerides to ensure accurate satellite positioning; 2) \emph{slice assembly}, mosaicking adjacent scenes acquired along the same orbital pass to ensure complete coverage of the study area; 3) \emph{radiometric calibration}, converting digital pixel values into normalized radar backscatter coefficients ($\sigma^0$); 4) \emph{Refined Lee speckle filtering}, dynamically adapting local filter windows based on local statistics to suppress multiplicative speckle noise while preserving parcel edges and texture details; 5) \emph{Range-Doppler terrain correction}, geocoding and radiometrically correcting geometric distortions such as layover and shadow using digital elevation model (DEM) data; 6) \emph{decibel (dB) conversion}, transforming linear backscatter coefficients to logarithmic scale via $\sigma^0_{\mathrm{dB}} = 10\log_{10}(\sigma^0)$; and 7) \emph{band computation}, deriving the polarimetric ratio VV/VH as the third backscatter channel.

\paragraph{Dual-Polarimetric $H/\alpha/A$ Decomposition} Using Single Look Complex (SLC) products as input, the pipeline sequentially executes precise orbit application, slice assembly, complex radiometric calibration, TOPSAR debursting, $\mathbf{C}_2$ polarimetric covariance matrix generation, spatial multilooking, dual-polarimetric $H/\alpha/A$ decomposition, and Range-Doppler terrain correction. Specifically, TOPSAR debursting removes dark burst-boundary demarcation zones and merges valid signals across the three Interferometric Wide-swath (IW) sub-swaths into a seamless image. Constructed from the dual-polarimetric scattering vector $\mathbf{k}=[S_{\mathrm{VV}}, S_{\mathrm{VH}}]^{\mathrm{T}}$, the $2\times 2$ polarimetric covariance matrix $\mathbf{C}_2=\langle\mathbf{k}\mathbf{k}^{\mathrm{H}}\rangle$ comprises two diagonal real-valued auto-power terms ($C_{11}=\langle|S_{\mathrm{VV}}|^2\rangle$ and $C_{22}=\langle|S_{\mathrm{VH}}|^2\rangle$) and one off-diagonal complex-valued cross-covariance term ($C_{12}=\langle S_{\mathrm{VV}}S_{\mathrm{VH}}^*\rangle$, which preserves inter-channel polarimetric coherence and phase difference). After spatial multilooking suppresses speckle noise and guarantees that $\mathbf{C}_2$ satisfies the non-singular full-rank condition ($\mathrm{rank}=2$), Hermitian eigen-decomposition extracts the dual-polarimetric entropy $H$, mean scattering angle $\alpha$, and anisotropy $A$.

\paragraph{Data Fusion, Resampling, and Normalization} The resulting backscatter channels (VV, VH, and VV/VH) and polarimetric decomposition channels ($H$, $\alpha$, and $A$) are resampled to a uniform 10\,m spatial resolution, reprojected to a common coordinate reference system, and stacked pixel-by-pixel to form six-band SAR time series. To eliminate scale discrepancies across physical quantities, per-band min--max normalization independently maps each of the six channels into the $[0, 1]$ range.

\section{Methodology}
\label{sec:method}

This section presents the proposed SAR-TS-BERT framework, organized into backscatter and polarimetric decomposition input design (Section~\ref{sec:input}), the dual-branch cross-modal fusion architecture (Section~\ref{sec:architecture}), the physics-guided span masking strategy (Section~\ref{sec:masking}), the pre-training and fine-tuning protocols (Section~\ref{sec:pretrain_finetune}), evaluation metrics (Section~\ref{sec:metrics}), and implementation details (Section~\ref{sec:implementation}).

\subsection{Backscatter and Polarimetric Decomposition Input Design}
\label{sec:input}

Let $\x \in \R^{T \times 6}$ denote a multitemporal SAR observation sequence comprising $T$ timesteps and six bands. Based on their physical scattering properties, the six input bands are partitioned into two functional groups:
\begin{equation}
\x_s = \x[:, 0\!:\!3] \in \R^{T \times 3}, \quad \x_p = \x[:, 3\!:\!6] \in \R^{T \times 3},
\end{equation}
where $\x_s$ represents the backscatter group (VV, VH, and VV/VH) and $\x_p$ denotes the dual-polarimetric Cloude--Pottier decomposition group ($H$, $\alpha$, and $A$). This partitioning reflects the intrinsic physical complementarity between radar echo intensity and polarimetric scattering structure. The model directly ingests pixel-level six-dimensional temporal sequences $\x \in \R^{T \times 6}$ as input (Fig.~\ref{fig:framework}(a)). Throughout this paper, $\x$ denotes the input sequence $X$ in Fig.~\ref{fig:framework}, and $\mathbf{H}$ denotes the encoder hidden representation to distinguish it from polarimetric entropy $H$.

\paragraph{Backscatter Bands} The co-polarized VV channel records radar returns sensitive to canopy surface geometry and soil moisture variations. The cross-polarized VH channel captures depolarized echoes primarily arising from multiple volume scattering within vegetation canopies. The polarimetric ratio VV/VH correlates strongly with canopy biomass~\cite{planque2021} and mitigates incidence-angle effects. Together, these three channels characterize the \emph{amplitude} response of radar echoes to crop geometry and dielectric water content.

\paragraph{Polarimetric Decomposition Bands} For Sentinel-1 dual-polarimetric (VV--VH) SLC acquisitions, we employ the dual-polarimetric Cloude--Pottier ($H/\alpha/A$) eigen-decomposition formulated on the $2\times 2$ polarimetric covariance matrix $\mathbf{C}_2$~\cite{harfenmeister2021,cloude_pottier1997}. Performing eigen-decomposition on the spatially multilooked $\mathbf{C}_2$ matrix yields
\begin{equation}
\mathbf{C}_2 = \sum_{i=1}^{2} \lambda_i \mathbf{u}_i \mathbf{u}_i^{\mathrm{H}}, \quad \lambda_1 \ge \lambda_2 \ge 0, \quad p_i = \frac{\lambda_i}{\lambda_1 + \lambda_2},
\end{equation}
where $\lambda_i$ denotes the non-negative real eigenvalues, $\mathbf{u}_i = [\cos\alpha_i, \sin\alpha_i e^{j\phi_i}]^{\mathrm{T}}$ represents the orthogonal unit eigenvectors, and $p_i$ is the pseudo-probability of the $i$-th scattering component. From these eigen-components, we derive polarimetric entropy $H = -\sum_{i=1}^{2} p_i \log_2 p_i$, mean scattering angle $\alpha = \sum_{i=1}^{2} p_i \alpha_i$, and anisotropy $A$. Within the VV--VH subspace, dual-polarimetric $\mathbf{C}_2$ decomposition characterizes target depolarization and polarimetric energy allocation: polarimetric entropy $H \in [0,1]$ measures scattering randomness on a logarithmic scale ($H \approx 0$ indicates deterministic scattering from bare soil or smooth water surfaces, whereas $H \approx 1$ corresponds to strong depolarized mixed scattering induced by dense crop canopies); mean scattering angle $\alpha \in [0^\circ, 90^\circ]$ reflects the projection ratio and phase coupling of the dominant scattering vector between co-polarized (VV) and cross-polarized (VH) channels, shifting systematically during stem elongation, heading, and canopy structural development; and anisotropy $A \in [0,1]$ characterizes the relative contrast between the secondary scattering component and the dominant polarimetric structure. Under the dual-polarimetric framework, jointly feeding $H$, $\alpha$, and $A$ into the neural network characterizes the underlying ground scattering \emph{mechanisms} across scattering randomness, dominant polarization angle, and polarimetric structure dimensions.

\paragraph{Physical Complementarity} The backscatter and polarimetric decomposition groups encode distinct physical properties: the former quantifies the total echo energy received by the sensor, whereas the latter reveals how that energy is distributed across scattering mechanisms. Crop types that exhibit similar temporal amplitude profiles often follow divergent scattering mechanism trajectories over the growing season, and vice versa. This physical complementarity motivates the dual-branch cross-modal architecture detailed below.

\subsection{SAR-TS-BERT Architecture}
\label{sec:architecture}

\begin{figure*}[t]
\centering
\includegraphics[width=0.90\textwidth]{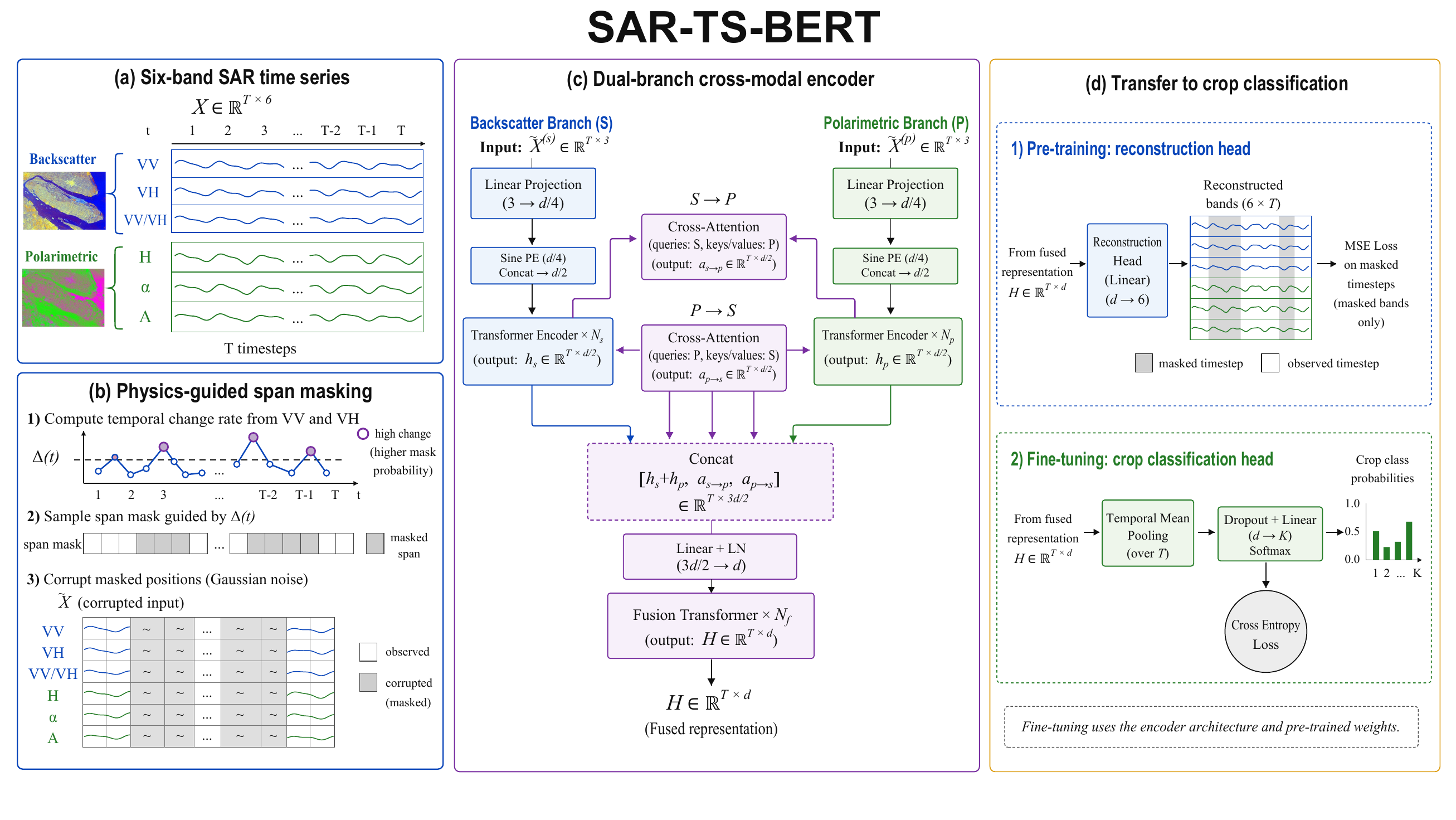}
\caption{Overall architecture of SAR-TS-BERT. (a) A six-band SAR time series comprises backscatter features (VV, VH, and VV/VH; blue) and polarimetric decomposition features ($H$, $\alpha$, and $A$; green), where image thumbnails and waveforms are illustrative. (b) Contiguous span sampling weights are computed from temporal backscatter change rates, and additive Gaussian noise is injected into all six bands at selected masked timesteps. (c) The two feature groups undergo separate linear projection, sinusoidal positional encoding concatenation, and branch-specific Transformer encoding, followed by bidirectional cross-attention and a fusion Transformer to generate temporal representations. (d) During pre-training, a linear reconstruction head restores the six-band observations and computes mean squared error (MSE) loss on masked timesteps; during fine-tuning, the pre-trained encoder is coupled with temporal mean pooling and a linear classification head trained with cross-entropy loss.}
\label{fig:framework}
\end{figure*}

Fig.~\ref{fig:framework} illustrates the overall architecture of SAR-TS-BERT. Self-supervised pre-training follows the reconstruction pathway along (a)$\to$(b)$\to$(c)$\to$(d), whereas supervised fine-tuning feeds uncorrupted six-band sequences directly into (c) and routes the fused representations to the classification branch in (d). In Fig.~\ref{fig:framework}(c), $\tilde{X}^{(s)}$ and $\tilde{X}^{(p)}$ denote the noise-corrupted inputs during pre-training, which correspond to the uncorrupted sequences $\x_s$ and $\x_p$ during fine-tuning.

The dual-branch cross-modal fusion encoder processes the backscatter group and the polarimetric decomposition group through dedicated Transformer branches before integrating them via bidirectional cross-attention.

\paragraph{Branch-Specific Encoding} Each branch first projects its three-dimensional input into a $d/4$-dimensional embedding via a linear layer, where $d$ is the model dimension:
\begin{equation}
\mathbf{e}_s = \text{Linear}_{s}(\x_s) \in \R^{T \times d/4}, \quad \mathbf{e}_p = \text{Linear}_{p}(\x_p) \in \R^{T \times d/4}.
\end{equation}
Sinusoidal positional encodings $\mathbf{p}_s, \mathbf{p}_p \in \R^{T \times d/4}$ are generated from the sequence timestep indices $t \in \{1,\ldots,T\}$ (shared across both branches, i.e., $\mathbf{p}_s = \mathbf{p}_p$) and concatenated with the projected embeddings along the feature dimension to form $d/2$-dimensional inputs, preventing low-dimensional physical features from being overwhelmed by additive positional signals in the initial layers:
\begin{equation}
\mathbf{z}_s = [\mathbf{e}_s \, \| \, \mathbf{p}_s] \in \R^{T \times d/2}, \quad \mathbf{z}_p = [\mathbf{e}_p \, \| \, \mathbf{p}_p] \in \R^{T \times d/2}.
\end{equation}
Each branch then passes through $N_s$ (backscatter branch) or $N_p$ (polarimetric branch) standard Transformer encoder layers~\cite{bert2019}, where each layer consists of multi-head self-attention (MHSA) and a feed-forward network (FFN) with residual connections and layer normalization:
\begin{align}
\mathbf{h}_s &= \text{TransformerEnc}_{s}(\mathbf{z}_s) \in \R^{T \times d/2}, \\
\mathbf{h}_p &= \text{TransformerEnc}_{p}(\mathbf{z}_p) \in \R^{T \times d/2}.
\end{align}

\paragraph{Cross-Modal Fusion} The two branch representations interact through bidirectional multi-head cross-attention. Denoting the interaction direction as $\text{query branch} \to \text{key/value branch}$ (where $s\to p$ indicates that $\mathbf{h}_s$ provides queries and $\mathbf{h}_p$ provides keys and values, and vice versa) and writing the multi-head cross-attention operator as $\operatorname{CrossAttn}(Q,K,V)$, we compute
\begin{align}
\mathbf{a}_{s \to p} &= \operatorname{CrossAttn}(\mathbf{h}_s, \mathbf{h}_p, \mathbf{h}_p), \\
\mathbf{a}_{p \to s} &= \operatorname{CrossAttn}(\mathbf{h}_p, \mathbf{h}_s, \mathbf{h}_s),
\end{align}
where $\mathbf{a}_{s\to p},\mathbf{a}_{p\to s}\in\R^{T\times d/2}$ are the bidirectional cross-modal attention output representations. Concatenating the element-wise sum of the branch outputs ($\mathbf{h}_s+\mathbf{h}_p$) with both cross-attention outputs along the feature dimension yields a $3d/2$-dimensional vector, which is projected back to dimension $d$ and normalized:
\begin{equation}
\mathbf{h}_f = \text{LayerNorm}\!\left(\text{Linear}\!\left([\mathbf{h}_s + \mathbf{h}_p \, \| \, \mathbf{a}_{s \to p} \, \| \, \mathbf{a}_{p \to s}]\right)\right) \in \R^{T \times d}.
\end{equation}
This formulation preserves both direct unimodal aggregation ($\mathbf{h}_s + \mathbf{h}_p$) and bidirectional cross-modal interactions ($\mathbf{a}_{s \to p}$ and $\mathbf{a}_{p \to s}$).

\paragraph{Fusion Encoder} The fused representation $\mathbf{h}_f$ is subsequently processed by $N_f = N_{\text{total}} - N_s - N_p$ Transformer encoder layers operating at the full model dimension $d$ to produce the final contextualized sequence representation $\mathbf{H} \in \R^{T \times d}$. We set $N_s=N_p=N_f=2$, resulting in $N_{\text{total}}=6$ Transformer encoder blocks in total; because the two front-end branches operate in parallel, each pathway from input to fused output traverses two branch-specific encoder layers followed by two shared fusion encoder layers.

\subsection{Physics-Guided Span Masking Strategy}
\label{sec:masking}

BERT-style pre-training relies on a masking strategy to select timesteps for reconstruction. Fig.~\ref{fig:framework}(b) illustrates the three sequential steps of our strategy: temporal change rate computation, contiguous span sampling, and Gaussian noise corruption.

We propose a contiguous span masking strategy driven by SAR backscatter temporal dynamics (SpanPhysics). Rapid changes in radar backscatter typically correspond to key phenological transitions—such as emergence, heading, and senescence—that carry the most discriminative information for crop identification. Masking contiguous spans around these high-change transitions compels the encoder to reconstruct critical phenological states from surrounding temporal context, yielding more discriminative dynamic representations.

Formally, given a SAR time series $\x \in \R^{T \times 6}$ ($T\geq2$), we compute the mean absolute first-order temporal difference across the VV and VH channels between adjacent timesteps ($t = 2, \ldots, T$). Because polarimetric decomposition parameters are more sensitive to residual local speckle perturbations, we compute temporal differences using the speckle-filtered VV and VH backscatter channels to robustly locate phenological transition boundaries (with 1-based timestep indexing and 0-based band indexing):
\begin{equation}
\Delta(t) = \frac{1}{2}\bigl(|\x[t, 0] - \x[t\!-\!1, 0]| + |\x[t, 1] - \x[t\!-\!1, 1]|\bigr).
\end{equation}
Subsequently, min--max normalization maps the sequence change rates to the $[0,1]$ interval. Letting $\Delta_{\min}$ and $\Delta_{\max}$ denote the minimum and maximum values of $\Delta(t)$ over $t=2,\ldots,T$, and setting $\varepsilon=10^{-6}$, the normalized change rate is defined as
\begin{equation}
r(t)=\begin{cases}
\displaystyle\frac{\Delta(t)-\Delta_{\min}}{\Delta_{\max}-\Delta_{\min}+\varepsilon},
&\Delta_{\max}>\Delta_{\min},\\[5pt]
0.5,&\Delta_{\max}=\Delta_{\min}.
\end{cases}
\label{eq:mask_rate_normalization}
\end{equation}
Truncating the configured span length to $1\leq L\leq T$ and letting $\ell=\lfloor L/2\rfloor$, each normalized change rate $r(u)$ accumulates weight toward candidate span starting positions within its local neighborhood:
\begin{equation}
\begin{split}
\mathcal{I}_u=\{s\in\{1,\ldots,T\}:{}\\
u-\ell\leq s\leq u+L-\ell-1\},
\end{split}
\end{equation}
for $u=2,\ldots,T$, where out-of-bounds indices are excluded by the set definition. The cumulative weight at candidate position $s$ is
\begin{equation}
S(s)=\sum_{u=2}^{T}r(u)\,\mathbf{1}\{s\in\mathcal{I}_u\},
\quad s=1,\ldots,T.
\end{equation}
The cumulative weights are scaled by the sequence maximum, shifted by a positive offset $\varepsilon$, and normalized over the set of valid starting positions $\mathcal{A}=\{1,\ldots,T-L+1\}$ that can accommodate a full span of length $L$:
\begin{align}
\bar{S}(s)&=\frac{S(s)}{\max_{1\leq v\leq T}S(v)}+\varepsilon,\\
p_{\mathrm{start}}(s)&=\frac{\bar{S}(s)}{\sum_{v\in\mathcal{A}}\bar{S}(v)},
\quad s\in\mathcal{A}.
\label{eq:mask_start_probability}
\end{align}
The positive offset $\varepsilon$ ensures a non-zero sampling probability for every valid starting position. The target number of masked timesteps and the nominal number of spans are given by
\begin{align}
k_{\mathrm{target}}&=\max(1,\lfloor T p_{\mathrm{mask}}\rfloor),\\
n_{\mathrm{spans}}&=\max(1,\lfloor k_{\mathrm{target}}/L\rfloor).
\end{align}
We sample $\min(n_{\mathrm{spans}},|\mathcal{A}|)$ starting positions $s$ without replacement according to $p_{\mathrm{start}}$ and take the union of the resulting contiguous segments $\{s,\ldots,s+L-1\}$ to form the masked index set $\mathcal{M}$. If $|\mathcal{M}|<k_{\mathrm{target}}$, additional timesteps are uniformly sampled without replacement from the unmasked positions until $|\mathcal{M}|=k_{\mathrm{target}}$; if $|\mathcal{M}|>2k_{\mathrm{target}}$, masked timesteps are uniformly removed without replacement until $|\mathcal{M}|=k_{\mathrm{target}}$; otherwise, $\mathcal{M}$ is retained as sampled. This sampling mechanism allows adjacent high-change segments to merge naturally, adapting the final masked set size $|\mathcal{M}|$ within $[k_{\mathrm{target}}, 2k_{\mathrm{target}}]$.

During the corruption stage, additive Gaussian noise is injected into all six bands at each masked timestep $t\in\mathcal{M}$: $\tilde{\x}[t] = \x[t] + \boldsymbol{\epsilon}_t$, where $\boldsymbol{\epsilon}_t \sim \mathcal{N}(\mathbf{0}, \sigma_n^2 \, \text{diag}(\boldsymbol{\sigma}_{\text{band}}^2))$ is sampled independently at each timestep. Here, $\sigma_n$ is the noise scale hyperparameter and $\boldsymbol{\sigma}_{\text{band}}$ denotes the per-band temporal standard deviation of the sample plus $10^{-6}$, while unmasked timesteps remain unchanged. Scaling soft Gaussian noise by the per-band temporal standard deviation preserves input distribution continuity, eliminates the representation discrepancy between pre-training and fine-tuning caused by discrete \texttt{[MASK]} tokens, and forces the encoder to suppress radiometric perturbations using surrounding temporal context to recover clean phenological trajectories.

\subsection{Pre-Training and Fine-Tuning}
\label{sec:pretrain_finetune}

\subsubsection{Self-Supervised Pre-Training}

Given an unlabeled SAR time series $\x \in \R^{T \times 6}$, the pre-training objective is to reconstruct the original six-band observations at the masked timesteps $\mathcal{M} \subset \{1, \ldots, T\}$ from the corrupted input $\tilde{\x}$. The dual-branch encoder maps the corrupted sequence to $\mathbf{H}\in\R^{T\times d}$, and a linear reconstruction head $\text{Linear}(d, 6)$ (top of Fig.~\ref{fig:framework}(d)) projects $\mathbf{H}$ back to the six-band observation space to produce predictions $\hat{\x}\in\R^{T\times6}$. The pre-training loss computes the mean squared error (MSE) strictly over the masked timesteps:
\begin{equation}
\mathcal{L}_{\text{pre}} = \frac{1}{|\mathcal{M}| \cdot 6} \sum_{t \in \mathcal{M}} \sum_{c=0}^{5} \left(\hat{\x}[t, c] - \x[t, c]\right)^2.
\end{equation}

\subsubsection{Supervised Fine-Tuning}

For downstream crop classification, we initialize the encoder (Fig.~\ref{fig:framework}(c)) with the pre-trained weights and replace the reconstruction head with a classification head (bottom of Fig.~\ref{fig:framework}(d)). Fine-tuning takes uncorrupted six-band sequences as input and jointly optimizes the encoder and low-rank adapters under a layer-wise learning rate schedule. The temporal representation $\mathbf{H} \in \R^{T \times d}$ undergoes temporal mean pooling across all $T$ timesteps, followed by dropout and a linear classification head with softmax activation to output class probabilities over $K$ crop categories:
\begin{equation}
\hat{y} = \text{Softmax}\!\left(\text{Linear}\!\left(\text{Dropout}\!\left(\frac{1}{T}\sum_{t=1}^{T} \mathbf{H}[t]\right)\right)\right).
\end{equation}
The model is optimized using the standard cross-entropy loss:
\begin{equation}
\mathcal{L}_{\text{cls}} = -\log \, \hat{y}[y].
\end{equation}

\paragraph{Parameter-Efficient Fine-Tuning via DoRA} To mitigate catastrophic forgetting of pre-trained physical representations, we adopt Weight-Decomposed Low-Rank Adaptation (DoRA)~\cite{dora2024}. DoRA decomposes each pre-trained weight matrix $\mathbf{W}_0 \in \R^{d_{\text{out}} \times d_{\text{in}}}$ into a magnitude vector $\mathbf{m}$ and a directional matrix $\mathbf{V}$, applying low-rank updates exclusively to the directional component:
\begin{equation}
\mathbf{W}' = \mathbf{m} \cdot \frac{\mathbf{V} + \Delta\mathbf{V}}{\|\mathbf{V} + \Delta\mathbf{V}\|_c}, \quad \Delta\mathbf{V} = \mathbf{B}\mathbf{A},
\end{equation}
where $\mathbf{A} \in \R^{r \times d_{\text{in}}}$ and $\mathbf{B} \in \R^{d_{\text{out}} \times r}$ are low-rank matrices with rank $r \ll \min(d_{\text{out}}, d_{\text{in}})$, and $\|\cdot\|_c$ denotes column-wise vector normalization. DoRA preserves parameter efficiency while approximating the learning capacity of full fine-tuning more closely than standard LoRA.

\paragraph{Layer-Wise Learning Rate Schedule} Differential learning rates are assigned across parameter groups during fine-tuning: the classification head uses the highest learning rate $\eta_{\text{head}}$, the DoRA adapters use an intermediate learning rate $\eta_{\text{dora}}$, and the pre-trained encoder backbone uses the lowest learning rate $\eta_{\text{enc}}$, satisfying $\eta_{\text{head}} > \eta_{\text{dora}} > \eta_{\text{enc}}$. Furthermore, the encoder backbone remains frozen during the first $w$ warmup epochs before being unfrozen under the layer-wise schedule.

\subsection{Evaluation Metrics}
\label{sec:metrics}

To evaluate classification performance across diverse crop distributions, we report Overall Accuracy ($\text{OA} = \sum_{k=1}^{K} C_{kk} / N$), Macro F1 ($\text{Macro F1} = \frac{1}{K}\sum_{k=1}^{K} \text{F1}_k$), and Weighted F1 ($\text{Weighted F1} = \sum_{k=1}^{K} \frac{n_k}{N} \text{F1}_k$), where $C_{ij}$ denotes the confusion matrix entry for ground-truth class $i$ and predicted class $j$, $K$ is the number of classes, $n_k$ is the test support of class $k$, $N$ is the total number of test samples, and $\text{F1}_k$ is the harmonic mean of precision and recall for class $k$. OA quantifies overall sample-level accuracy, Macro F1 weights all classes equally to reflect sensitivity to minority crops, and Weighted F1 accounts for class imbalance. We also report per-class F1 scores and the pre-training gain ($\Delta\text{F1}$) relative to training from scratch.

\subsection{Implementation Details}
\label{sec:implementation}

Table~\ref{tab:implementation_details} summarizes the hyperparameter configurations for pre-training, fine-tuning, and model architecture in SAR-TS-BERT.

\begin{table}[t]
\centering
\caption{Implementation Details of SAR-TS-BERT}
\label{tab:implementation_details}
\resizebox{\linewidth}{!}{%
\begin{tabular}{lll}
\toprule
Stage / Component & Parameter & Setting \\
\midrule
\multirow{7}{*}{Pre-training}
& Epochs & 100 \\
& Batch size & 512 \\
& Optimizer & AdamW \\
& Learning rate & $1\times10^{-4}$ \\
& Masking ratio & $p_{\text{mask}}=0.15$ \\
& Span length & $L=3$ \\
& Noise scale & $\sigma_n=0.1$ \\
\midrule
\multirow{8}{*}{Fine-tuning}
& Epochs & 300 \\
& Early stopping patience & 35 \\
& Batch size & 256 \\
& Classification head LR & $\eta_{\text{head}}=1\times10^{-3}$ \\
& DoRA adapter LR & $\eta_{\text{dora}}=5\times10^{-4}$ \\
& Encoder backbone LR & $\eta_{\text{enc}}=1\times10^{-4}$ \\
& DoRA rank & $r=32$ \\
& Encoder warmup epochs & $w=10$ \\
\midrule
\multirow{5}{*}{Model Architecture}
& Model dimension & $d_{\text{model}}=256$ \\
& Attention heads & $n_{\text{head}}=8$ \\
& Encoder layers & $n_{\text{layers}}=6$ (2 backscatter + 2 polarimetric + 2 fusion) \\
& Feed-forward dimension & $\text{dim}_{\text{ff}}=512$ \\
& Total parameters & $\sim$3.4M \\
\bottomrule
\end{tabular}%
}
\end{table}

For baseline methods, Random Forest (RF) is implemented in scikit-learn with 100 decision trees, while BiRNN and BiLSTM are implemented in PyTorch as two-layer bidirectional networks with a hidden dimension of 256. All baseline models ingest the same six-band SAR time-series inputs. For external self-supervised baselines (SITS-BERT-SAR and Presto-SAR), each model first executes its native self-supervised pre-training objective on the unlabeled six-band SAR data before supervised fine-tuning; the remaining supervised baselines (PSE-TAE-SAR, ExViT-TS, and TSViT-SAR) are trained end-to-end under full supervision.

\section{Experiments}
\label{sec:experiments}

\subsection{Experiments on the Estonia Dataset}
\label{sec:estonia_results}

We first evaluate classification accuracy and parcel-level mapping reliability on the Estonia dataset using the six-band SAR time series (VV, VH, VV/VH, and $H/\alpha/A$). For baseline architectures originally designed for optical or multi-source time series, we adapt their input projection layers to six-band SAR sequences while keeping their core temporal encoding modules and optimal hyperparameters intact; external self-supervised baselines (SITS-BERT-SAR and Presto-SAR) are pre-trained on the unlabeled six-band SAR data prior to fine-tuning. Tables~\ref{tab:estonia_main} and~\ref{tab:estonia_perclass} report the overall and per-class comparison results on the Estonia test set.

\begin{table}[!t]
\centering
\caption{Main Quantitative Comparison on the Estonia Dataset}
\label{tab:estonia_main}
\resizebox{\linewidth}{!}{%
\begin{tabular}{lcccc}
\toprule
Method & Params & OA & Macro F1 & Weighted F1 \\
\midrule
Random Forest & - & 0.8577 & 0.7946 & 0.8429 \\
BiRNN (2-layer) & 403K & 0.8987 & 0.8628 & 0.8976 \\
BiLSTM (2-layer) & 537K & 0.9023 & 0.8681 & 0.9002 \\
\midrule
SITS-BERT-SAR~\cite{sitsbert2021} & 3.20M & 0.9501 & 0.9335 & 0.9498 \\
PSE-TAE-SAR~\cite{garnot2020pse} & 3.46M & 0.9427 & 0.9239 & 0.9427 \\
ExViT-TS~\cite{exvit2023} & 3.17M & 0.9316 & 0.9107 & 0.9316 \\
Presto-SAR~\cite{presto2023} & 3.20M & 0.9276 & 0.9048 & 0.9273 \\
TSViT-SAR~\cite{tsvit2023} & 4.83M & 0.9263 & 0.9017 & 0.9259 \\
\midrule
SAR-TS-BERT (From Scratch) & $\sim$3.4M & 0.9561 & 0.9413 & 0.9557 \\
\textbf{SAR-TS-BERT (With Pre-training)} & \textbf{$\sim$3.4M} & \textbf{0.9678} & \textbf{0.9572} & \textbf{0.9677} \\
\bottomrule
\end{tabular}%
}
\end{table}

\begin{table}[!t]
\centering
\caption{Per-Class F1 Score Comparison on the Estonia Dataset}
\label{tab:estonia_perclass}
\resizebox{\linewidth}{!}{%
\begin{tabular}{lcccc}
\toprule
Crop Type & F1 (From Scratch) & F1 (With Pre-training) & $\Delta$ F1 & Test Samples \\
\midrule
Clover & 0.9044 & 0.9366 & +3.22\% & 15,704 \\
Legumes harvested green & 0.8895 & 0.9205 & +3.10\% & 15,516 \\
Oats & 0.9293 & 0.9518 & +2.25\% & 12,255 \\
Pasture/meadow/grassland & 0.9710 & 0.9795 & +0.85\% & 129,742 \\
Peas & 0.9607 & 0.9693 & +0.86\% & 15,233 \\
Spring barley & 0.9488 & 0.9595 & +1.07\% & 23,080 \\
Spring wheat & 0.9276 & 0.9484 & +2.08\% & 11,012 \\
Winter wheat & 0.9650 & 0.9723 & +0.73\% & 35,416 \\
Winter rapeseed & 0.9758 & 0.9765 & +0.07\% & 12,142 \\
\midrule
\textbf{Macro Avg} & \textbf{0.9413} & \textbf{0.9572} & \textbf{+1.59\%} & 270,100 \\
\bottomrule
\end{tabular}%
}
\end{table}

\paragraph{Key Classification Results on Estonia} Pre-trained SAR-TS-BERT achieves the highest overall accuracy (0.9678) and Macro F1 (0.9572), improving OA and Macro F1 by 1.17 and 1.59 percentage points over training from scratch and outperforming conventional classifiers (RF, BiRNN, and BiLSTM) by at least 6.55 percentage points in OA. Among adapted SOTA models, SAR-TS-BERT surpasses the strongest external baseline (SITS-BERT-SAR) by 1.77 and 2.37 percentage points in OA and Macro F1, respectively, and exceeds the remaining Transformer baselines by 2.51--4.15 percentage points in OA. While the competitive performance of SITS-BERT-SAR confirms the suitability of BERT-style masked reconstruction for SAR time series, SAR-TS-BERT further advances accuracy at a comparable parameter scale via dual-branch polarimetric encoding, SpanPhysics masking, and DoRA fine-tuning. Pre-training improves F1 across all nine crop types (Table~\ref{tab:estonia_perclass}), with the largest gains on clover (+3.22 percentage points) and legumes harvested green (+3.10 percentage points), demonstrating enhanced separability among crops with similar backscatter profiles.

\begin{figure*}[!t]
\centering
\includegraphics[width=0.90\textwidth]{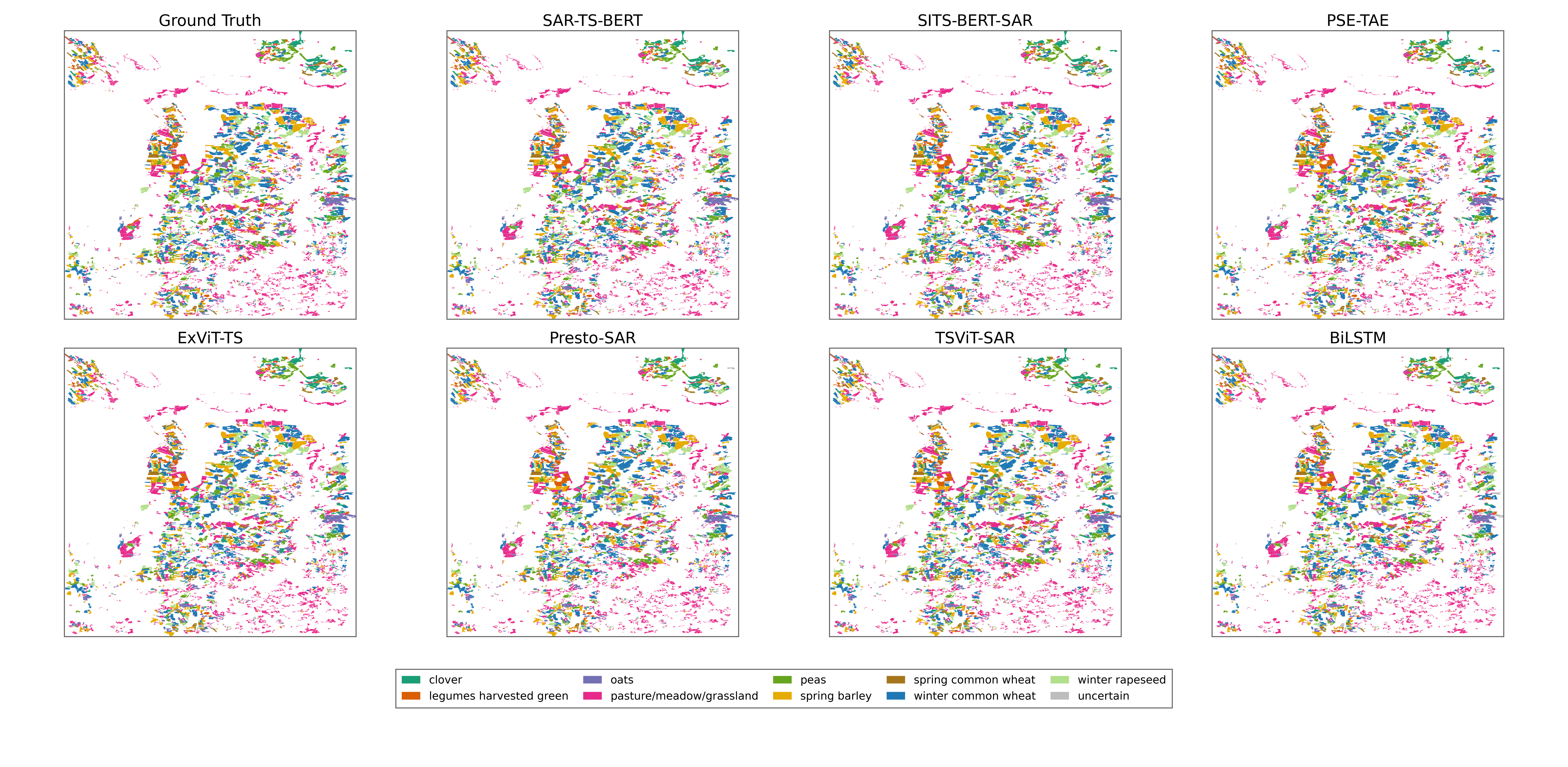}
\caption{Comparison of parcel-level crop prediction maps on the Estonia test set.}
\label{fig:estonia_threshold06_prediction}
\end{figure*}

\begin{figure*}[!t]
\centering
\includegraphics[width=0.90\textwidth]{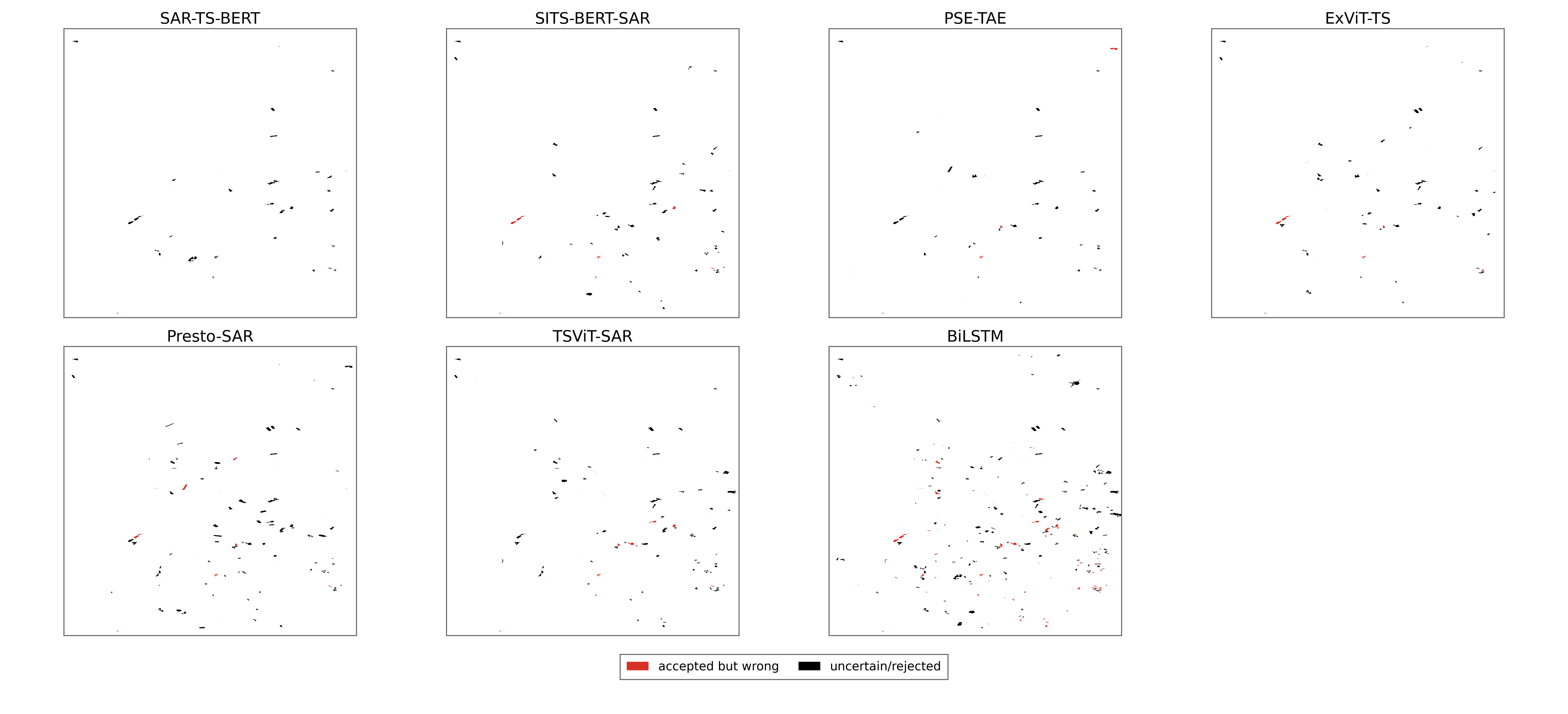}
\caption{Spatial distribution of accepted wrong parcels and uncertain/rejected parcels on the Estonia test set. Red indicates parcels accepted under the voting threshold ($r_p \ge 0.6$) but misclassified; black indicates low-confidence rejected parcels ($r_p < 0.6$).}
\label{fig:estonia_threshold06_error}
\end{figure*}

\paragraph{Parcel-Level Visualization and Mapping Reliability Analysis} To evaluate spatial consistency in parcel-level mapping, we aggregate pixel-level predictions across all 2,832 independent test parcels via majority voting. For parcel $p$ with $n_p$ test pixels and $n_{p,c}$ pixels predicted as class $c$, let the dominant-class voting ratio be
\begin{equation}
r_p=\frac{n_{p,c^\ast}}{n_p}, \quad c^\ast=\arg\max_c n_{p,c}.
\end{equation}
Parcels with $r_p < 0.6$ are flagged as low-confidence (\emph{uncertain/rejected}) rather than forced into a crop class, jointly quantifying classification accuracy and within-parcel spatial agreement. Figs.~\ref{fig:estonia_threshold06_prediction} and~\ref{fig:estonia_threshold06_error} and Table~\ref{tab:estonia_threshold06_mapping} present the parcel-level mapping results.

\begin{table*}[!t]
\centering
\caption{Threshold-Constrained Parcel-Level Mapping Results on the Estonia Test Set (Total Test Parcels: 2,832)}
\label{tab:estonia_threshold06_mapping}
\resizebox{\textwidth}{!}{%
\begin{tabular}{lcccccc}
\toprule
Method & Full Parcel OA & Accepted OA & Accepted Macro F1 & Coverage & Rejected Parcels & Accepted Wrong Parcels \\
\midrule
\textbf{SAR-TS-BERT} & \textbf{0.9958} & \textbf{0.9982} & \textbf{0.9980} & \textbf{98.69\%} & \textbf{37} & \textbf{5} \\
PSE-TAE-SAR & 0.9944 & 0.9964 & 0.9958 & 98.38\% & 46 & 10 \\
SITS-BERT-SAR & 0.9905 & 0.9946 & 0.9939 & 98.09\% & 54 & 15 \\
ExViT-TS & 0.9901 & 0.9935 & 0.9927 & 98.13\% & 53 & 18 \\
Presto-SAR & 0.9862 & 0.9931 & 0.9923 & 96.61\% & 96 & 19 \\
TSViT-SAR & 0.9848 & 0.9931 & 0.9921 & 96.93\% & 87 & 19 \\
BiLSTM & 0.9516 & 0.9744 & 0.9655 & 92.51\% & 212 & 67 \\
\bottomrule
\end{tabular}%
}
\end{table*}

SAR-TS-BERT achieves the highest Full Parcel OA (0.9958) and Accepted OA (0.9982) while maintaining the highest parcel coverage (98.69\%), producing only 37 rejected parcels and 5 accepted wrong parcels across all 2,832 test parcels—halving the accepted errors of the runner-up PSE-TAE-SAR (10 wrong, 46 rejected). By contrast, Presto-SAR and TSViT-SAR require rejecting 96 and 87 parcels to reach an Accepted OA of 0.9931, and BiLSTM drops to 92.51\% coverage with 67 accepted wrong parcels. As shown in Figs.~\ref{fig:estonia_threshold06_prediction} and~\ref{fig:estonia_threshold06_error}, competing models suffer from concentrated rejections and misclassifications among structurally similar forage and legume crops (legumes harvested green, clover, and pasture/meadow/grassland), whereas SAR-TS-BERT yields fewer and more dispersed errors, confirming superior within-parcel prediction consistency.

\subsection{Experiments on the USA Dataset}
\label{sec:usa_results}

We next evaluate generalization on the large-scale pixel-level USA (California CDL) dataset, where the held-out 75\% independent test split comprises 21,712,842 pixel samples across eight crop types. Tables~\ref{tab:usa_main} and~\ref{tab:usa_perclass} summarize the overall and per-class results on the independent test set.

\begin{table}[!t]
\centering
\caption{Main Quantitative Comparison on the USA Dataset}
\label{tab:usa_main}
\resizebox{\linewidth}{!}{%
\begin{tabular}{lcccc}
\toprule
Method & Params & OA & Macro F1 & Weighted F1 \\
\midrule
Random Forest & - & 0.8232 & 0.7832 & 0.8191 \\
BiRNN (2-layer) & 403K & 0.8182 & 0.7656 & 0.8157 \\
BiLSTM (2-layer) & 537K & 0.8200 & 0.7761 & 0.8169 \\
\midrule
SITS-BERT-SAR~\cite{sitsbert2021} & 3.20M & 0.8689 & 0.8547 & 0.8663 \\
ExViT-TS~\cite{exvit2023} & 3.17M & 0.8613 & 0.8458 & 0.8606 \\
TSViT-SAR~\cite{tsvit2023} & 4.83M & 0.8605 & 0.8437 & 0.8595 \\
PSE-TAE-SAR~\cite{garnot2020pse} & 3.46M & 0.8603 & 0.8433 & 0.8589 \\
Presto-SAR~\cite{presto2023} & 3.20M & 0.8561 & 0.8382 & 0.8550 \\
\midrule
SAR-TS-BERT (From Scratch) & $\sim$3.4M & 0.8503 & 0.8274 & 0.8490 \\
\textbf{SAR-TS-BERT (With Pre-training)} & \textbf{$\sim$3.4M} & \textbf{0.8729} & \textbf{0.8616} & \textbf{0.8723} \\
\bottomrule
\end{tabular}%
}
\end{table}

\begin{table}[!t]
\centering
\caption{Per-Class F1 Score Comparison on the USA Dataset}
\label{tab:usa_perclass}
\resizebox{\linewidth}{!}{%
\begin{tabular}{lcccc}
\toprule
Crop Type & F1 (From Scratch) & F1 (With Pre-training) & $\Delta$ F1 & Test Samples \\
\midrule
Corn & 0.9358 & 0.9463 & +1.05\% & 2,350,136 \\
Winter Wheat & 0.9558 & 0.9660 & +1.02\% & 1,589,918 \\
Alfalfa & 0.8542 & 0.8836 & +2.94\% & 1,644,062 \\
Tomatoes & 0.9446 & 0.9531 & +0.85\% & 1,642,029 \\
Grapes & 0.8406 & 0.8635 & +2.29\% & 7,139,181 \\
Almonds & 0.7667 & 0.8014 & +3.47\% & 4,489,549 \\
Walnuts & 0.8295 & 0.8519 & +2.24\% & 2,699,423 \\
Pistachios & 0.4923 & 0.6269 & \textbf{+13.46\%} & 158,544 \\
\midrule
\textbf{Macro Avg} & \textbf{0.8274} & \textbf{0.8616} & +3.42\% & 21,712,842 \\
\bottomrule
\end{tabular}%
}
\end{table}

\paragraph{Key Classification Results on the USA Dataset} Affected by boundary mixed pixels and CDL label noise, shallow recurrent models (BiRNN and BiLSTM) perform slightly below Random Forest (0.8232 OA), while supervised Transformer baselines (ExViT-TS, TSViT-SAR, and PSE-TAE-SAR) cluster tightly around 0.8603--0.8613 OA. Pre-trained SAR-TS-BERT achieves the best overall performance (0.8729 OA and 0.8616 Macro F1), improving OA and Macro F1 by 2.26 and 3.42 percentage points over training from scratch and outperforming the runner-up SITS-BERT-SAR by 0.40 and 0.69 percentage points. Per-class comparisons (Table~\ref{tab:usa_perclass}) show that pre-training yields the largest F1 gains on minority and perennial orchard classes, notably pistachios (+13.46 percentage points), almonds (+3.47 percentage points), and alfalfa (+2.94 percentage points).

\begin{figure*}[!t]
\centering
\includegraphics[width=0.90\textwidth]{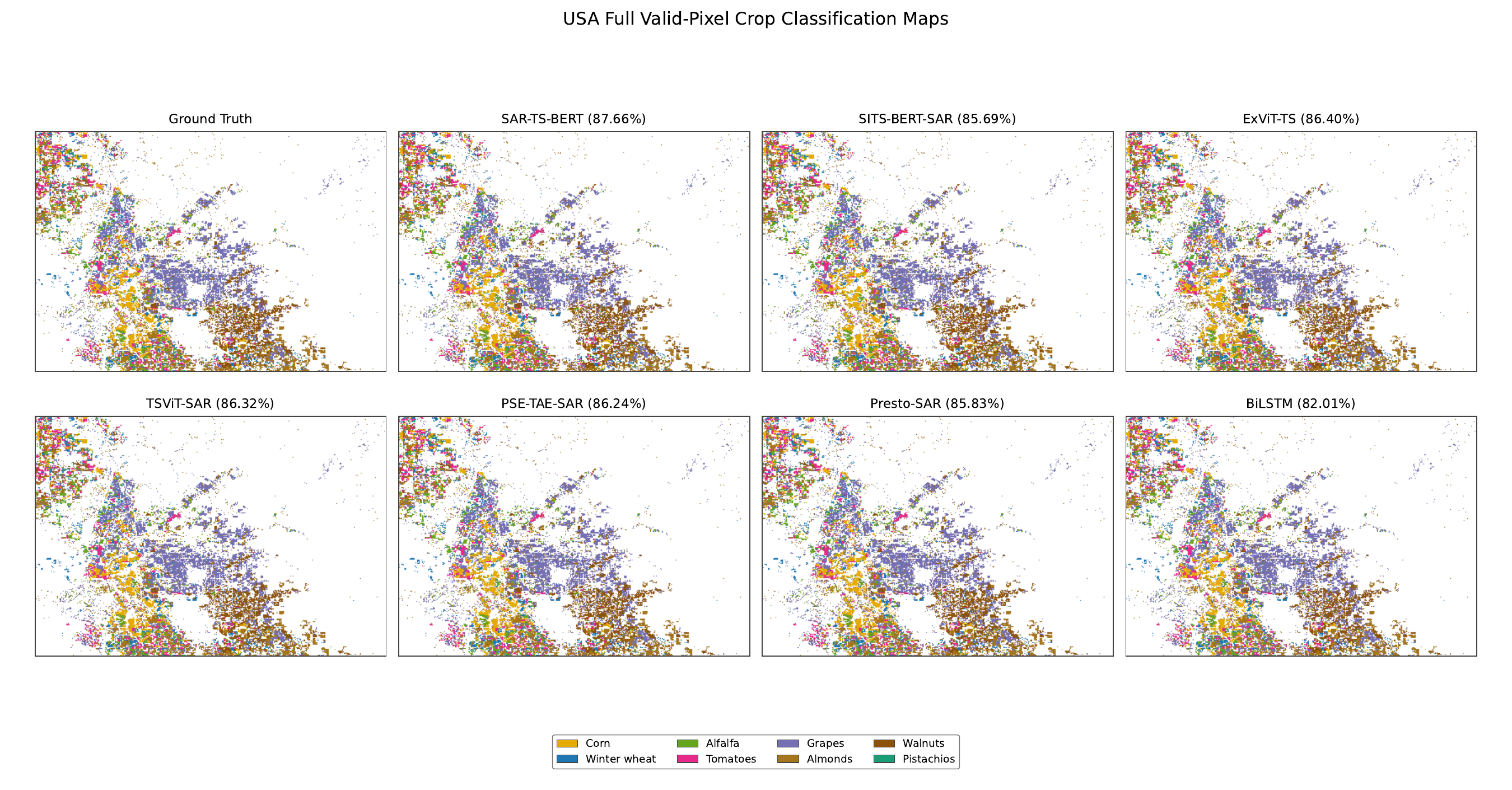}
\caption{Comparison of full valid-pixel spatial crop classification maps on the USA dataset. All models use the six-band SAR time-series input and are rendered under the same CDL color legend.}
\label{fig:usa_sota_prediction}
\end{figure*}

\begin{figure*}[!t]
\centering
\includegraphics[width=0.90\textwidth]{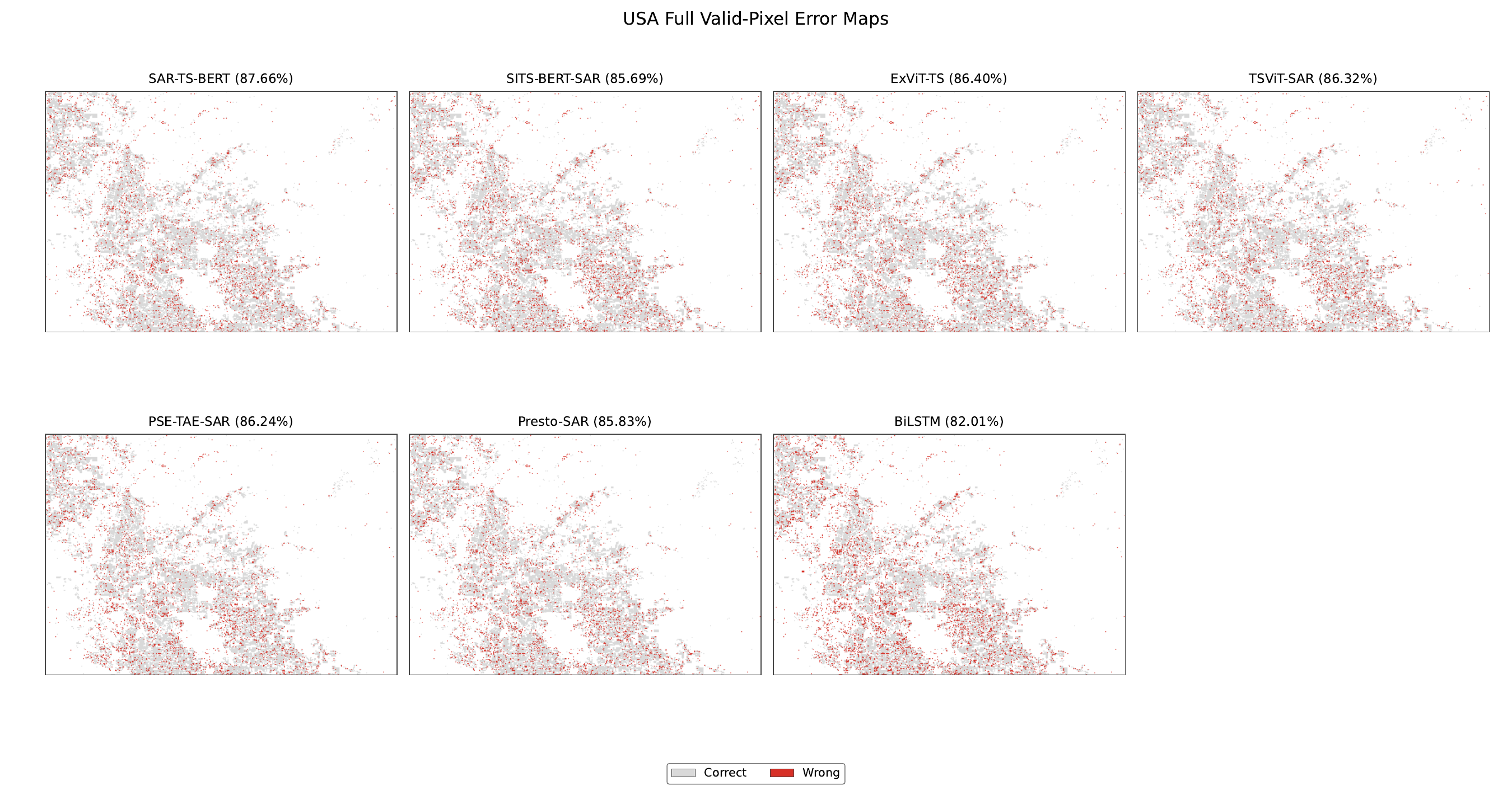}
\caption{Comparison of full valid-pixel error distributions on the USA dataset. Gray denotes pixels matching the CDL reference labels, and red denotes misclassified pixels.}
\label{fig:usa_sota_error}
\end{figure*}

\paragraph{Full Valid-Pixel Spatial Mapping on the USA Dataset} Distinct from Table~\ref{tab:usa_main} (which evaluates generalization strictly on the held-out 21.71M independent test split), we further conduct wall-to-wall inference across all 28,950,454 valid pixels in the entire USA study scene to inspect continuous raster mapping quality, field-boundary coherence, and spatial error clustering (Table~\ref{tab:usa_full_valid_mapping} and Figs.~\ref{fig:usa_sota_prediction}--\ref{fig:usa_sota_error}).

\begin{table}[!t]
\centering
\caption{Full Valid-Pixel Spatial Mapping Accuracy on the USA Dataset (Six-Band Input, 28,950,454 Valid Pixels)}
\label{tab:usa_full_valid_mapping}
\resizebox{\linewidth}{!}{%
\begin{tabular}{lccc}
\toprule
Method & OA & Macro F1 & Weighted F1 \\
\midrule
\textbf{SAR-TS-BERT} & \textbf{0.8766} & \textbf{0.8663} & \textbf{0.8760} \\
SITS-BERT-SAR & 0.8569 & 0.8399 & 0.8563 \\
ExViT-TS & 0.8640 & 0.8492 & 0.8634 \\
TSViT-SAR & 0.8632 & 0.8472 & 0.8622 \\
PSE-TAE-SAR & 0.8624 & 0.8462 & 0.8610 \\
Presto-SAR & 0.8583 & 0.8413 & 0.8572 \\
BiLSTM & 0.8201 & 0.7764 & 0.8171 \\
\bottomrule
\end{tabular}%
}
\end{table}

Across the complete valid-pixel scene, SAR-TS-BERT achieves the highest OA (0.8766) and Macro F1 (0.8663), surpassing the strongest baseline (ExViT-TS) by 1.26 and 1.71 percentage points. Visually, SAR-TS-BERT preserves cleaner field coherence across contiguous corn, winter wheat, and tomato parcels while reducing speckle-like misclassifications inside almond, walnut, and pistachio orchards (Fig.~\ref{fig:usa_sota_error}), confining residual errors primarily to field boundaries and intercropped orchard margins.

\subsection{Experiments on Chongming Island}
\label{sec:chongming}

Representing a typical double-cropping region, Chongming Island is evaluated across independent binary rice/wheat identification and 11-class monthly land-cover classification.

\subsubsection{Binary Identification of Rice and Wheat}

As shown in Table~\ref{tab:chongming_binary}, self-supervised pre-training consistently improves both independent binary tasks, raising OA and Macro F1 by 0.19 and 0.38 percentage points for rice identification (reaching 0.9963 OA and 0.9921 Macro F1), and by 0.18 and 0.19 percentage points for wheat identification (reaching 0.9923 OA and 0.9921 Macro F1).

\begin{table}[t]
\centering
\caption{Independent Binary Classification Results for Rice and Wheat on Chongming Island}
\label{tab:chongming_binary}
\begin{tabular}{llcc}
\toprule
Task & Method & OA & Macro F1 \\
\midrule
\multirow{2}{*}{Rice} & From Scratch & 0.9944 & 0.9883 \\
 & \textbf{With Pre-training} & \textbf{0.9963} & \textbf{0.9921} \\
\midrule
\multirow{2}{*}{Wheat} & From Scratch & 0.9905 & 0.9902 \\
 & \textbf{With Pre-training} & \textbf{0.9923} & \textbf{0.9921} \\
\bottomrule
\end{tabular}
\end{table}

\subsubsection{Monthly Crop and Land-Cover Classification}

\begin{figure}[t]
\centering
\includegraphics[width=\linewidth]{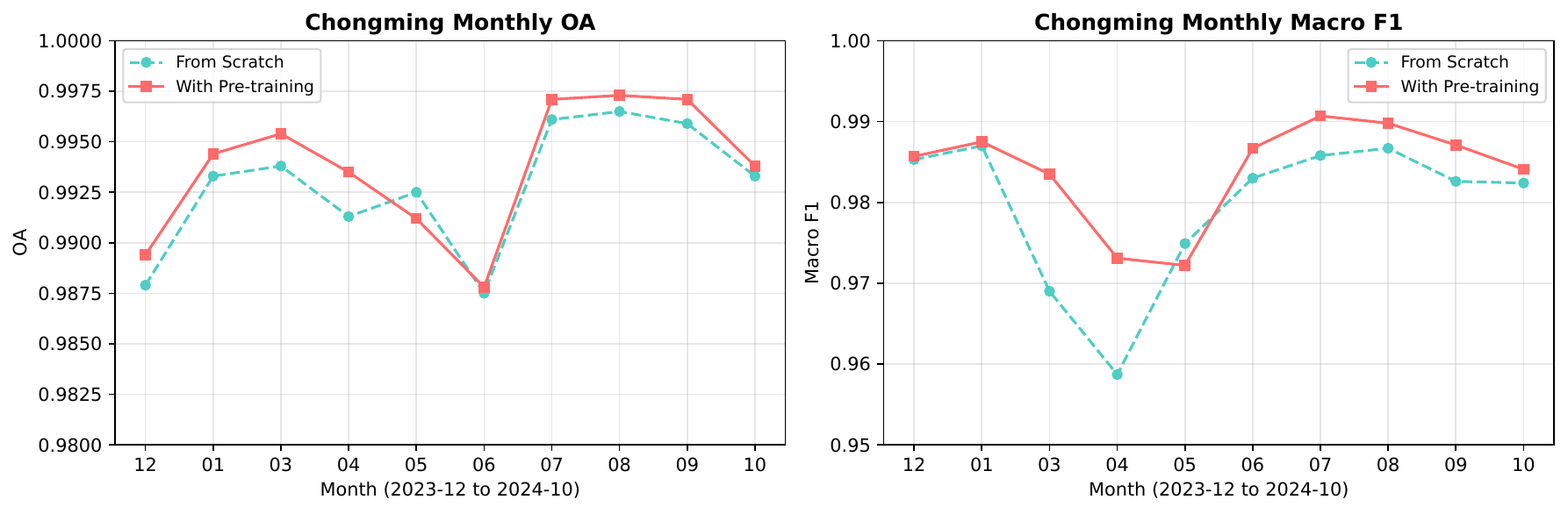}
\caption{Monthly crop classification comparison on Chongming Island (11 classes). Left: Overall Accuracy (OA); right: Macro F1. The pre-trained model achieves higher OA in 9 out of 10 months, with pronounced Macro F1 gains during the March--April phenological transition period.}
\label{fig:chongming_monthly}
\end{figure}

Fig.~\ref{fig:chongming_monthly} and Table~\ref{tab:chongming_perclass} present the 11-class monthly classification results across the 10 valid annotated months from December 2023 to October 2024 (225,970 pooled test samples).

\begin{table}[t]
\centering
\caption{Per-Class F1 Scores on Chongming Island (Pooled Across 10 Months, 11 Classes, 225,970 Samples)}
\label{tab:chongming_perclass}
\resizebox{\linewidth}{!}{%
\begin{tabular}{lcccc}
\toprule
Category & F1 (From Scratch) & F1 (With Pre-training) & $\Delta$ F1 & Test Samples \\
\midrule
Rice & 0.9967 & 0.9970 & +0.03\% & 90,979 \\
Wheat & 0.9958 & 0.9962 & +0.04\% & 52,824 \\
Corn & 0.9909 & 0.9914 & +0.05\% & 3,570 \\
Soybean & 0.9812 & 0.9873 & +0.61\% & 5,503 \\
Vegetables & 0.9840 & 0.9813 & -0.27\% & 2,206 \\
Rapeseed & 0.9886 & 0.9886 & 0.00\% & 5,952 \\
Woodland & 0.9764 & 0.9803 & +0.39\% & 8,870 \\
Citrus & 0.9760 & 0.9879 & \textbf{+1.19\%} & 2,330 \\
Rice stubble & 0.9900 & 0.9910 & +0.10\% & 16,031 \\
Plowed land & 0.9883 & 0.9891 & +0.08\% & 34,897 \\
Weeds & 0.9886 & 0.9929 & +0.43\% & 2,808 \\
\midrule
\textbf{Macro Avg} & \textbf{0.9870} & \textbf{0.9894} & \textbf{+0.24\%} & 225,970 \\
\bottomrule
\end{tabular}%
}
\end{table}

\paragraph{Key Findings} Pre-training improves monthly OA in 9 out of 10 months (raising mean monthly OA from 0.9928 to 0.9937) and increases pooled Macro F1 by 0.24 percentage points (from 0.9870 to 0.9894). Gains are most pronounced during the March--April spring phenological transition window when land-cover composition is most heterogeneous, boosting Macro F1 by 1.45 percentage points in March (0.9690 to 0.9835) and 1.44 percentage points in April (0.9587 to 0.9731). At the class level, minority categories such as citrus (+1.19 percentage points) and soybean (+0.61 percentage points) benefit most from pre-training.

\subsection{Ablation Study: Six-Band Fusion vs.\ Three-Band Inputs}
\label{sec:ablation_6band}

\begin{figure}[t]
\centering
\includegraphics[width=\linewidth]{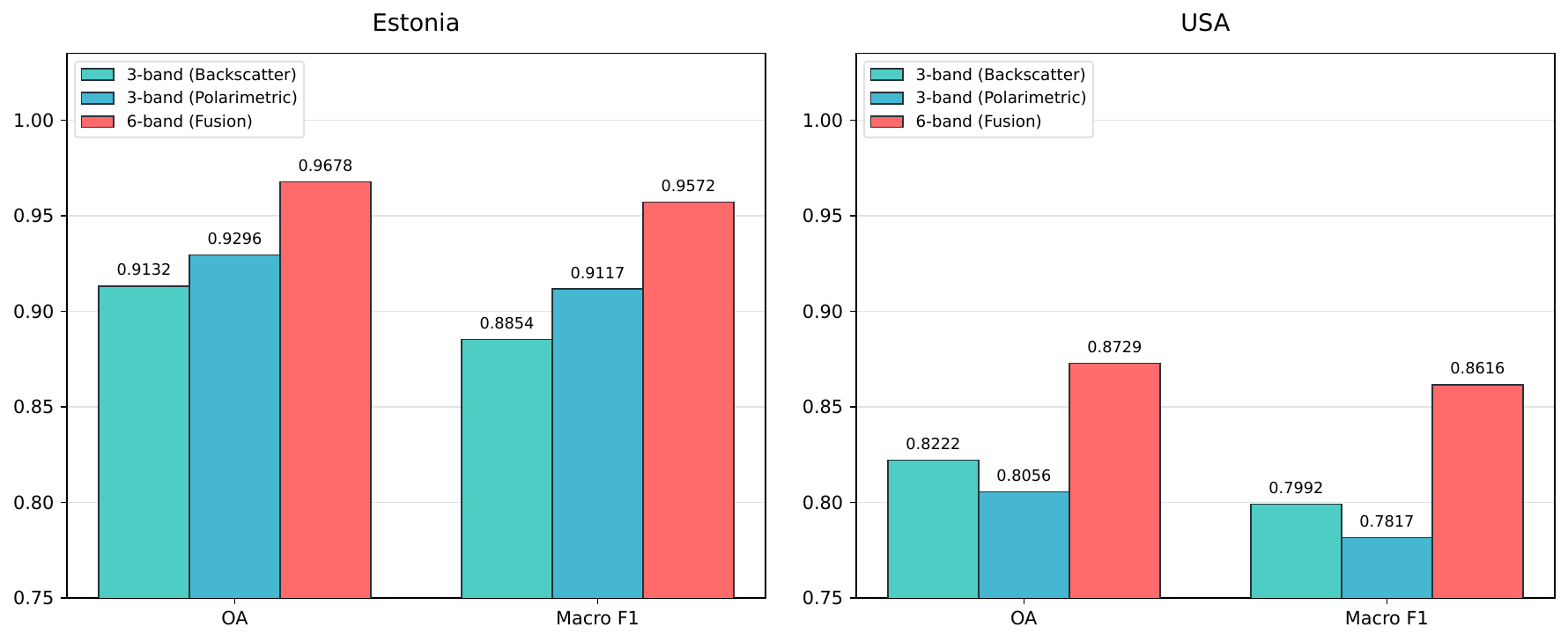}
\caption{Ablation comparison of 6-band fusion versus 3-band inputs. Left: Estonia dataset; right: USA dataset. Comparing 3-band (Backscatter), 3-band (Polarimetric), and 6-band (Fusion) confirms strong physical complementarity between the two feature groups.}
\label{fig:6band_vs_3band}
\end{figure}

To quantify the contribution of fusing dual-polarimetric decomposition features ($H/\alpha/A$) with backscatter intensities, we compare three input configurations under the exact same pre-training and fine-tuning pipeline: 1) \emph{3-band (Backscatter)} (VV, VH, and VV/VH), capturing echo amplitude alone; 2) \emph{3-band (Polarimetric)} ($H$, $\alpha$, and $A$), capturing scattering mechanisms alone; and 3) \emph{6-band (Fusion)} (VV, VH, VV/VH, $H$, $\alpha$, and $A$), integrating both groups. All three configurations are evaluated on the independent test sets of Estonia and the USA (21,712,842 test samples, matching Table~\ref{tab:usa_main}). Fig.~\ref{fig:6band_vs_3band} and Table~\ref{tab:6band_ablation} present the results.

\begin{table}[t]
\centering
\caption{Ablation Study on 6-Band vs.\ 3-Band Inputs (Pre-Trained Models). 3-Band (Backscatter): VV, VH, and VV/VH; 3-Band (Polarimetric): $H/\alpha/A$.}
\label{tab:6band_ablation}
\begin{tabular}{llccc}
\toprule
Dataset & Input Configuration & OA & Macro F1 & Weighted F1 \\
\midrule
\multirow{3}{*}{Estonia} & 3-band (Backscatter) & 0.9132 & 0.8854 & 0.9129 \\
 & 3-band (Polarimetric) & 0.9296 & 0.9117 & 0.9273 \\
 & \textbf{6-band (Fusion)} & \textbf{0.9678} & \textbf{0.9572} & \textbf{0.9677} \\
\midrule
\multirow{3}{*}{USA} & 3-band (Backscatter) & 0.8222 & 0.7992 & 0.8204 \\
 & 3-band (Polarimetric) & 0.8056 & 0.7817 & 0.8026 \\
 & \textbf{6-band (Fusion)} & \textbf{0.8729} & \textbf{0.8616} & \textbf{0.8723} \\
\bottomrule
\end{tabular}
\end{table}

\paragraph{Key Findings} Table~\ref{tab:6band_ablation} highlights three findings: 1) \emph{Six-band fusion consistently achieves the highest accuracy}, improving OA on Estonia (0.9678) by 5.46 and 3.82 percentage points over the backscatter-only and polarimetric-only inputs, and on the USA (0.8729) by 5.07 and 6.73 percentage points, respectively; 2) \emph{The relative strength of the two three-band subsets varies by region}—polarimetric features lead on Estonia (0.9296 vs.\ 0.9132 OA) due to structural scattering differences among spring grains, legumes, and grasslands, whereas backscatter features lead on the USA (0.8222 vs.\ 0.8056 OA) due to strong amplitude contrasts across irrigated crops and orchards, yet fusing both groups yields substantial gains in both regimes; and 3) on Estonia, even the six-band model trained from scratch (0.9561 OA, Table~\ref{tab:estonia_main}) outperforms the pre-trained three-band backscatter model (0.9132 OA) by 4.29 percentage points, confirming the fundamental value of joint backscatter--polarimetric modeling.

\subsection{Ablation Study: Dual-Branch Encoding and Cross-Attention}
\label{sec:ablation_architecture}

To isolate the contributions of grouped dual-branch encoding and explicit bidirectional cross-attention, we fix the six-band input $\x=[\mathrm{VV},\mathrm{VH},\mathrm{VV/VH},H,\alpha,A]$ and compare three architectural variants:
\begin{itemize}
    \item \textbf{Variant A (Single-Concat)}: Projects all six bands jointly to 128 dimensions, concatenates a 128-dimensional positional encoding, and processes the 256-dimensional sequence through two single-branch Transformer layers, a $256\to 256$ linear projection and layer normalization module, and two subsequent Transformer layers.
    \item \textbf{Variant B (Dual-NoCross)}: Encodes the backscatter and polarimetric groups separately via two-layer Transformer branches to obtain $\mathbf{h}_s,\mathbf{h}_p\in\R^{T\times128}$, fuses them via static linear projection without cross-attention:
    \begin{equation}
    \mathbf{F}_B=\operatorname{LN}\!\left(W_f[\mathbf{h}_s+\mathbf{h}_p\,\|\,\mathbf{h}_s\,\|\,\mathbf{h}_p]+b_f\right),
    \label{eq:ablation_static_fusion}
    \end{equation}
    where $W_f$ maps the 384-dimensional concatenation to 256 dimensions, and passes $\mathbf{F}_B$ through two shared fusion Transformer layers.
    \item \textbf{Variant C (Dual-Cross)}: Our full architecture, which retains the same dual-branch front-end encoder as Variant B and replaces the raw branch concatenation with bidirectional cross-attention outputs $[\mathbf{h}_s+\mathbf{h}_p\,\|\,\mathbf{a}_{s\to p}\,\|\,\mathbf{a}_{p\to s}]\in\R^{T\times 384}$ before the $384\to 256$ fusion projection and two-layer fusion encoder. Thus, Variants B and C share identical input and output dimensions ($384\to 256$) at the fusion projection layer.
\end{itemize}

In all three variants, every pathway from input to fused output traverses four Transformer encoder layers (two front-end layers followed by two subsequent layers). All variants use identical six-band inputs, preprocessing pipelines, dataset splits, and unlabeled pre-training samples, undergoing independent pre-training (SpanPhysics masking with $p_{\text{mask}}=0.15$, $L=3$, and $\sigma_n=0.1$) and fine-tuning under identical optimization budgets, learning rate schedules, DoRA configurations, validation selection rules, and output heads on the Estonia and USA independent test sets (21,712,842 test samples on the USA dataset, matching Table~\ref{tab:usa_main}).

\begin{table}[t]
\centering
\caption{Ablation Results of Dual-Branch Encoding and Cross-Attention Under Fixed Six-Band Input. A: Single-Concat; B: Dual-NoCross; C: Dual-Cross.}
\label{tab:architecture_ablation}
\begin{tabular}{llccc}
\toprule
Dataset & Variant & OA & Macro F1 & Weighted F1 \\
\midrule
\multirow{3}{*}{Estonia} & A & 0.9537 & 0.9363 & 0.9531 \\
 & B & 0.9615 & 0.9488 & 0.9612 \\
 & C & \textbf{0.9678} & \textbf{0.9572} & \textbf{0.9677} \\
\midrule
\multirow{3}{*}{USA} & A & 0.8604 & 0.8471 & 0.8601 \\
 & B & 0.8676 & 0.8568 & 0.8669 \\
 & C & \textbf{0.8729} & \textbf{0.8616} & \textbf{0.8723} \\
\bottomrule
\end{tabular}
\end{table}

\paragraph{Key Findings} As shown in Table~\ref{tab:architecture_ablation}, decoupled dual-branch encoding (Variant B) improves OA and Macro F1 over early concatenation (Variant A) by 0.78 and 1.25 percentage points on Estonia, and by 0.72 and 0.97 percentage points on the USA dataset. Incorporating explicit bidirectional cross-attention (Variant C) yields further gains of 0.63 and 0.84 percentage points in OA and Macro F1 on Estonia, and 0.53 and 0.48 percentage points on the USA (reaching total Macro F1 improvements of 2.09 and 1.45 percentage points over Variant A), validating both modality-specific branch encoding and cross-branch attention interaction.

\subsection{Ablation Study: Masking Strategies}
\label{sec:ablation_masking}

\begin{figure}[t]
\centering
\includegraphics[width=\linewidth]{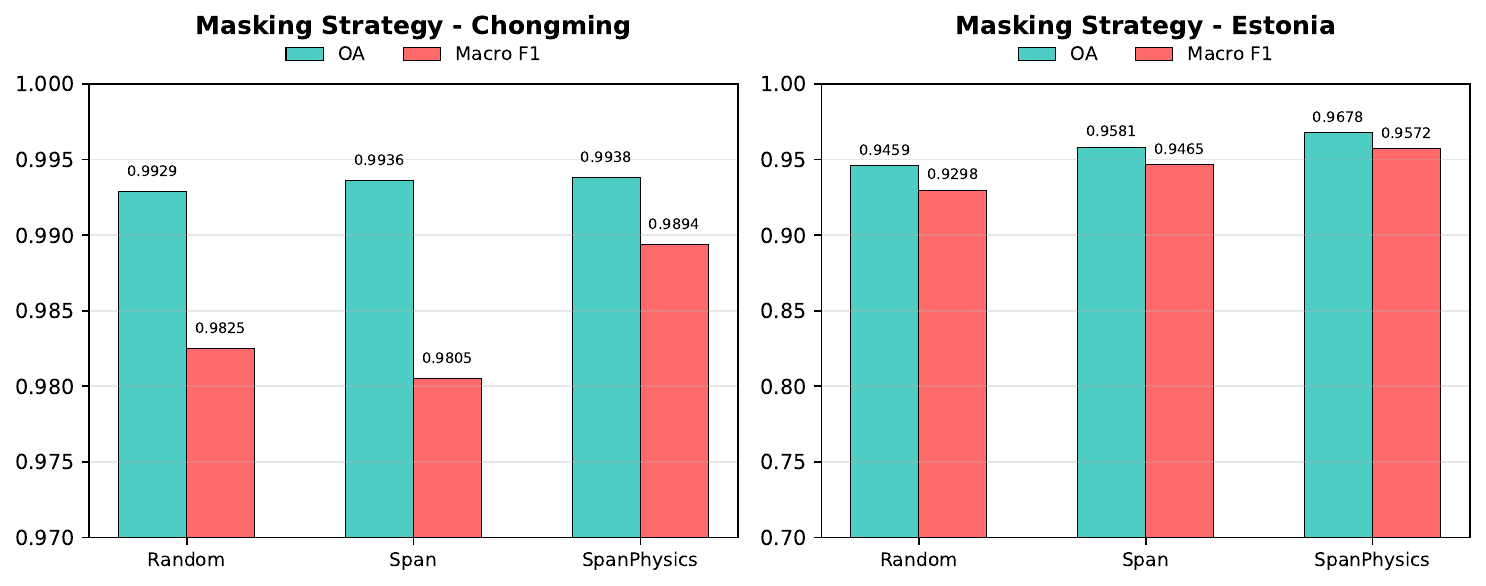}
\caption{Ablation comparison of masking strategies. Left: Chongming Island dataset; right: Estonia dataset. SpanPhysics achieves the highest OA and Macro F1 on both benchmarks.}
\label{fig:masking_ablation}
\end{figure}

We compare our physics-guided contiguous span masking (\textbf{SpanPhysics}, where span starting probabilities follow temporal backscatter change rates) against uniform random timestep masking (\textbf{Random}, as in SITS-BERT~\cite{sitsbert2021}) and uniform contiguous span masking (\textbf{Span}, with random starting positions and $L=3$) at $p_{\text{mask}}=0.15$. Fig.~\ref{fig:masking_ablation} and Table~\ref{tab:masking_ablation} summarize the results on Chongming Island and Estonia.

\begin{table}[t]
\centering
\caption{Ablation Study on Masking Strategies (Masking Ratio $p_{\text{mask}}=0.15$, Span Length $L=3$)}
\label{tab:masking_ablation}
\begin{tabular}{llccc}
\toprule
Dataset & Masking Strategy & OA & Macro F1 \\
\midrule
\multirow{3}{*}{Chongming} & Random & 0.9929 & 0.9825 \\
 & Span (len=3) & 0.9936 & 0.9805 \\
 & \textbf{SpanPhysics} & \textbf{0.9938} & \textbf{0.9894} \\
\midrule
\multirow{3}{*}{Estonia} & Random & 0.9459 & 0.9298 \\
 & Span (len=3) & 0.9581 & 0.9465 \\
 & \textbf{SpanPhysics} & \textbf{0.9678} & \textbf{0.9572} \\
\bottomrule
\end{tabular}
\end{table}

\paragraph{Key Findings} SpanPhysics achieves the highest OA and Macro F1 on both benchmarks, improving Macro F1 by 0.69 and 0.89 percentage points over Random and Span masking on Chongming Island, and improving OA/Macro F1 on Estonia by 2.19/2.74 percentage points over Random masking and by 0.97/1.07 percentage points over uniform Span masking. While unguided Span masking slightly trails Random masking in Macro F1 on Chongming Island, anchoring contiguous spans to rapid backscatter transitions consistently strengthens phenological representation learning.

\subsection{Few-Shot Evaluation}
\label{sec:fewshot}

\begin{figure}[t]
\centering
\includegraphics[width=\linewidth]{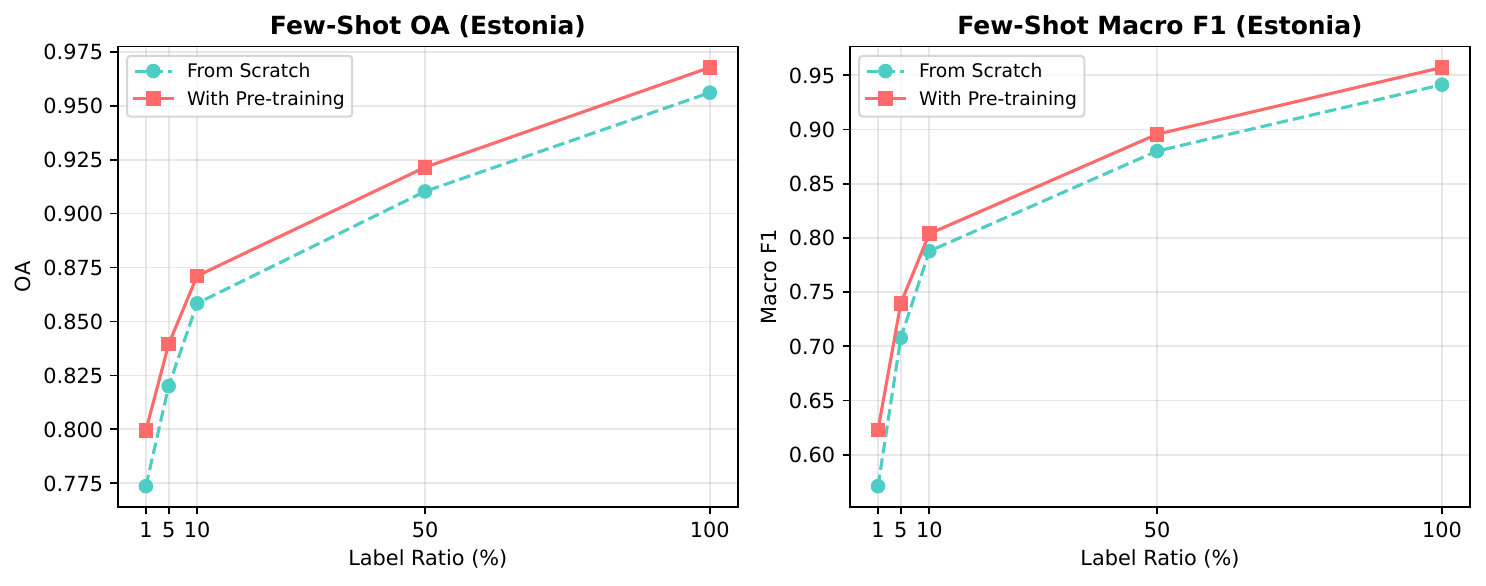}
\caption{Few-shot evaluation results on the Estonia dataset. Left: Overall Accuracy (OA); right: Macro F1. Pre-training consistently outperforms training from scratch across all label ratios, with amplified gains under the 1\% and 5\% label regimes.}
\label{fig:fewshot}
\end{figure}

To evaluate self-supervised pre-training under label scarcity, we fine-tune SAR-TS-BERT using randomly sampled 1\%, 5\%, 10\%, 50\%, and 100\% subsets of the Estonia labeled training data while evaluating on the fixed 50\% test set. Specifically, the pre-trained model is first pre-trained for 100 epochs on the complete unlabeled Estonia dataset using SpanPhysics masking before supervised fine-tuning under the standard protocol (Table~\ref{tab:implementation_details}), whereas the from-scratch baseline uses the exact same fine-tuning configuration without loading pre-trained weights. Fig.~\ref{fig:fewshot} and Table~\ref{tab:fewshot} compare the two regimes.

\begin{table}[t]
\centering
\caption{Few-Shot Evaluation Results on the Estonia Dataset}
\label{tab:fewshot}
\resizebox{\linewidth}{!}{%
\begin{tabular}{lcccccc}
\toprule
\multirow{2}{*}{Label Ratio} & \multicolumn{2}{c}{OA} & \multirow{2}{*}{$\Delta$ OA} & \multicolumn{2}{c}{Macro F1} & \multirow{2}{*}{$\Delta$ F1} \\
\cmidrule(lr){2-3} \cmidrule(lr){5-6}
 & From Scratch & With Pre-training & & From Scratch & With Pre-training & \\
\midrule
1\% & 0.7736 & 0.7994 & \textbf{+2.58\%} & 0.5710 & 0.6229 & \textbf{+5.19\%} \\
5\% & 0.8200 & 0.8395 & \textbf{+1.95\%} & 0.7078 & 0.7397 & \textbf{+3.19\%} \\
10\% & 0.8584 & 0.8711 & +1.27\% & 0.7877 & 0.8038 & +1.61\% \\
50\% & 0.9103 & 0.9215 & +1.12\% & 0.8801 & 0.8956 & +1.55\% \\
100\% & 0.9561 & 0.9678 & +1.17\% & 0.9413 & 0.9572 & +1.59\% \\
\bottomrule
\end{tabular}%
}
\end{table}

\paragraph{Key Findings} Pre-training consistently outperforms training from scratch across all label budgets and exhibits a pronounced \emph{few-shot gain amplification} effect: improvements peak at 1\% labels ($\Delta\text{OA}=+2.58$ and $\Delta\text{F1}=+5.19$ percentage points) and 5\% labels ($\Delta\text{OA}=+1.95$ and $\Delta\text{F1}=+3.19$ percentage points), while remaining stable at +1.55 to +1.59 percentage points in Macro F1 under 50\%--100\% supervision, confirming that self-supervised pre-training provides robust parameter initialization when ground-truth annotations are scarce.

\section{Discussion}
\label{sec:discussion}

Moving beyond metric comparisons, this section examines the physical mechanisms and operational boundaries of SAR-TS-BERT across three dimensions: 1) the physical complementarity between backscatter intensities and polarimetric decomposition parameters; 2) how this complementarity governs classification and spatial mapping reliability across different annotation granularities; and 3) remaining methodological limitations and future directions.

\subsection{SAR Temporal Responses and Polarimetric Feature Complementarity}
\label{sec:discussion_feature_complementarity}

The rice--wheat double-cropping system on Chongming Island provides a physically grounded testbed for interpreting the six-band SAR representation. Because the same agricultural parcels undergo a winter--spring wheat season followed by a summer--autumn rice season within a single annual cycle, they experience continuous transitions in crop species, canopy architecture, surface water conditions, and soil roughness. Using the 25 Sentinel-1 acquisitions from December 2023 to October 2024, we analyze the six-band temporal profiles for fixed rice parcels, fixed wheat parcels, and exact wheat--rice rotation parcels (excluding February 2024 from class-wise aggregation due to the absence of monthly ground-truth labels).

\begin{figure*}[t]
\centering
\includegraphics[width=0.90\textwidth]{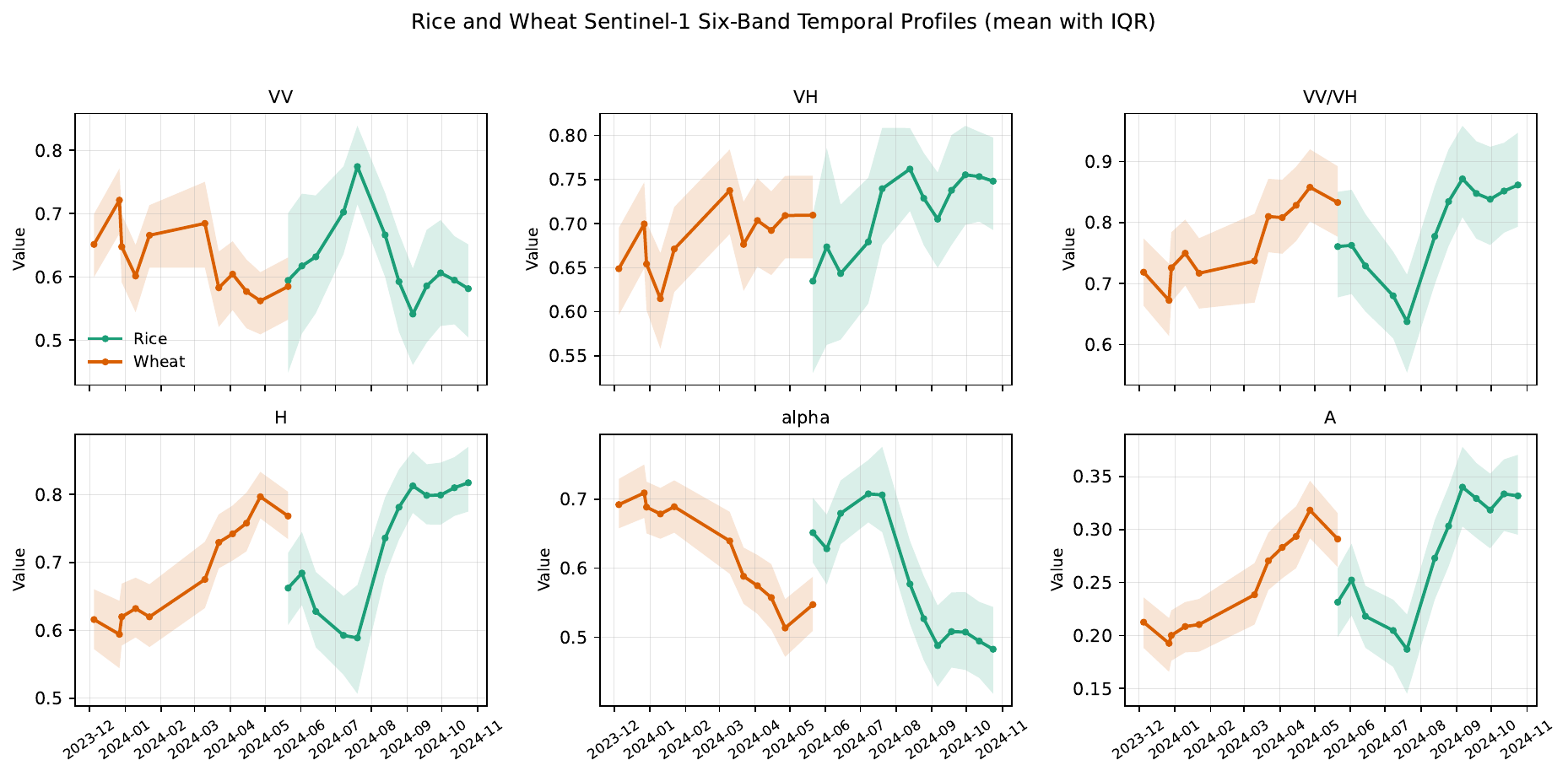}
\caption{Six-band Sentinel-1 temporal profiles for rice and wheat on Chongming Island. Solid curves denote class means, and shaded bands indicate interquartile ranges (IQR).}
\label{fig:discussion_exp31_sixband}
\end{figure*}

As shown in Fig.~\ref{fig:discussion_exp31_sixband}, both rice and wheat exhibit pronounced seasonality across VV, VH, VV/VH, and $H/\alpha/A$, yet their peak windows and trajectory directions differ substantially across feature groups. Backscatter features primarily reflect echo amplitude variations governed jointly by canopy volume scattering, soil background, and moisture conditions; for instance, wheat exhibits strong VH and VV/VH responses from stem elongation through heading, whereas rice displays a distinct temporal phase shift from flooding and transplanting through tillering and maturity. In contrast, $H/\alpha/A$ characterizes variations in scattering randomness, mean scattering angle, and polarimetric anisotropy. During the late growth stages of wheat (March--May), $H$ and $A$ generally increase while $\alpha$ decreases, reflecting canopy densification and enhanced vegetation--ground mixed scattering. For rice, $\alpha$ remains elevated while $H$ and $A$ stay at a trough from transplanting to early--mid tillering (June--July); subsequently, from booting to late maturity (August--October), $H$ and $A$ increase substantially while $\alpha$ declines markedly, consistent with the attenuation of the flooded water background, evolving plant geometry, and pre-harvest surface state changes. These patterns demonstrate that polarimetric decomposition parameters provide temporal scattering information complementary to backscatter amplitude.

\begin{figure*}[t]
\centering
\includegraphics[width=0.90\textwidth]{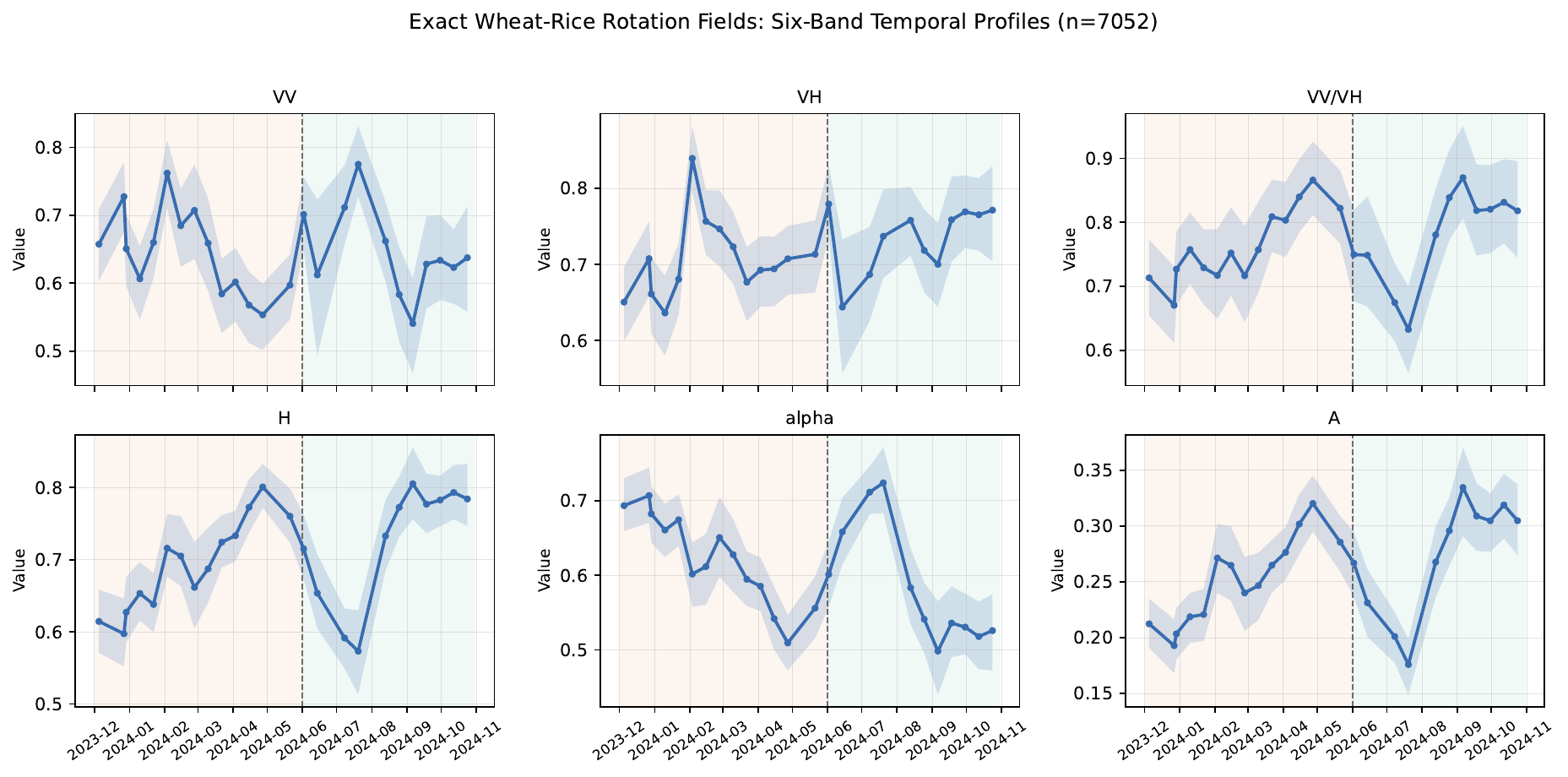}
\caption{Temporal evolution of the six SAR features within the exact same parcels transitioning from winter--spring wheat to summer--autumn rice.}
\label{fig:discussion_exp31_rotation}
\end{figure*}

Fig.~\ref{fig:discussion_exp31_rotation} further corroborates this mechanism by tracking continuous intra-parcel transitions across exact wheat--rice rotation fields, reducing confounding variations from soil texture and field management. As parcels transition from the wheat season into the rice season, VV, VH, and VV/VH trace the amplitude-driven phenological profile, whereas $H$, $\alpha$, and $A$ exhibit sharper regime shifts across the seasonal turnover, supplying essential discriminative cues when amplitude trajectories converge or change gradually.

To quantify this complementarity, we compute the absolute Cohen's $d$ as a univariate class separability metric for acquisition date $t$, feature band $b$, and crop classes $c_1$ and $c_2$:
\begin{equation}
d_{t,b}=
\frac{\left|\mu_{1,t,b}-\mu_{2,t,b}\right|}
{\left[
\frac{(n_1-1)\sigma_{1,t,b}^{2}+(n_2-1)\sigma_{2,t,b}^{2}}
{n_1+n_2-2}
\right]^{1/2}},
\label{eq:cohens_d_discussion}
\end{equation}
where $\mu$, $\sigma$, and $n$ denote the class sample mean, standard deviation, and sample size on date $t$ and band $b$, respectively. We also compute the group-averaged separability for the backscatter and polarimetric decomposition groups:
\begin{equation}
\bar{d}_{\mathrm{back}}=\frac{d_{\mathrm{VV}}+d_{\mathrm{VH}}+d_{\mathrm{VV/VH}}}{3},
\quad
\bar{d}_{\mathrm{pol}}=\frac{d_{H}+d_{\alpha}+d_{A}}{3}.
\label{eq:group_d_discussion}
\end{equation}

\begin{figure*}[t]
\centering
\includegraphics[width=0.90\textwidth]{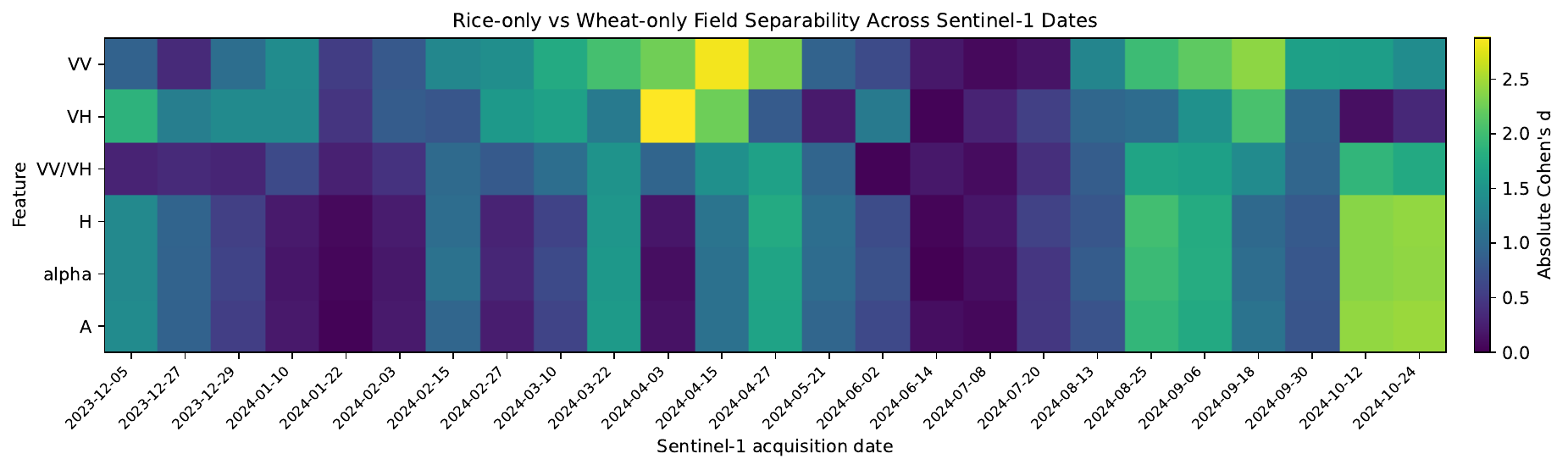}
\caption{Temporal separability heatmap (absolute Cohen's $d$) across the six SAR bands between fixed wheat parcels and fixed rice parcels on Chongming Island. Higher values indicate stronger inter-class separability on a given acquisition date and feature.}
\label{fig:discussion_fixedfield_heatmap}
\end{figure*}

Fig.~\ref{fig:discussion_fixedfield_heatmap} shows that the high-separability windows of backscatter and polarimetric decomposition features are temporally asynchronous. Peak separability for VH and VV occurs around April 2024, capturing the amplitude contrast between rapidly growing winter wheat canopies and fallow/flooded rice paddies, whereas peak separability ($|d|$) for $A$, $H$, and $\alpha$ concentrates around October 2024, reflecting heightened sensitivity to late-maturity rice canopy structure. Combining both feature groups therefore expands the effective temporal window for crop differentiation.

\begin{table}[t]
\centering
\caption{Six-Band Separability (Absolute Cohen's $d$) Between Rice and Wheat Across Analytical Settings on Chongming Island}
\label{tab:discussion_separability_summary}
\resizebox{\linewidth}{!}{%
\begin{tabular}{lccc}
\toprule
Feature & Same-Month Rice vs.\ Wheat & Pooled Wheat vs.\ Rice Seasons & Exact Wheat--Rice Rotation Parcels \\
\midrule
VV & 0.073 & 0.039 & 0.205 \\
VH & 0.629 & 0.426 & 0.571 \\
VV/VH & 0.686 & 0.153 & 0.070 \\
$H$ & 1.674 & 0.399 & 0.347 \\
$\alpha$ & 1.659 & 0.472 & 0.387 \\
$A$ & 1.456 & 0.481 & 0.399 \\
\bottomrule
\end{tabular}%
}
\end{table}

Table~\ref{tab:discussion_separability_summary} further confirms that in the strict same-month rice-versus-wheat comparison, the separability of $H$, $\alpha$, and $A$ substantially exceeds that of VV, VH, and VV/VH, whereas in the pooled seasonal comparison and exact rotation parcels, VH and $H/\alpha/A$ jointly maintain strong separability. This asynchronous physical response directly explains the six-band fusion gains in Section~\ref{sec:ablation_6band} and motivates the grouped dual-branch cross-attention architecture validated in Section~\ref{sec:ablation_architecture}.

\subsection{Cross-Dataset Performance and Spatial Mapping Reliability}
\label{sec:discussion_cross_dataset_mapping}

Comparing performance across the Estonia, USA, and Chongming Island benchmarks reveals how annotation granularity and landscape structure shape classification accuracy and error distributions. On Estonia, parcel-level sampling and homogeneous field interiors mitigate single-pixel speckle perturbations and boundary mixing, enabling SAR-TS-BERT to learn clean phenological trajectories (0.9678 OA). By contrast, the USA dataset relies on pixel-level CDL raster annotations across California's Central Valley, where perennial orchards (almonds, walnuts, and pistachios) share similar canopy architectures, row planting geometries, and irrigation regimes, while boundary mixed pixels and inherent CDL label noise cap the achievable pixel-level ceiling on the held-out independent test split (0.8729 OA, Table~\ref{tab:usa_main}). Crucially, self-supervised pre-training yields its largest per-class F1 gains on these challenging minority and perennial categories (Table~\ref{tab:usa_perclass}), demonstrating that pre-trained temporal--polarimetric representations effectively regularize fine-tuning under label noise and class imbalance.

In operational spatial mapping, decoupled physical branch encoding and cross-modal attention translate pixel-level discrimination into superior intra-parcel agreement and raster continuity. Under confidence thresholding on Estonia (Table~\ref{tab:estonia_threshold06_mapping}), competing models must reject 46--212 parcels to maintain high accepted accuracy, whereas SAR-TS-BERT achieves both the highest Accepted OA (0.9982) and highest coverage (98.69\%) with only 37 rejected and 5 wrong parcels. Complementing the held-out test evaluation in Table~\ref{tab:usa_main}, the wall-to-wall inference experiment across all 28.95 million valid pixels in the USA scene (Table~\ref{tab:usa_full_valid_mapping} and Fig.~\ref{fig:usa_sota_error}) evaluates continuous raster mapping quality and spatial error clustering, confirming that SAR-TS-BERT suppresses interior field fragmentation and confines residual misclassifications primarily to parcel borders and heterogeneous orchard margins where single-season C-band SAR faces physical discriminability limits.

Furthermore, the monthly evaluation on Chongming Island (Fig.~\ref{fig:chongming_monthly}) demonstrates that pre-training gains concentrate during the March--April spring phenological turnover. Aligning with our Cohen's $d$ separability analysis, rapid vegetative growth, harvest, and rotation turnover induce concurrent shifts in backscatter amplitude and polarimetric scattering mechanisms. By prioritizing these high-change transition segments during SpanPhysics pre-training, SAR-TS-BERT maintains stable discrimination across rapid phenological transitions, low-label regimes, and spatially heterogeneous agricultural landscapes.

\subsection{Limitations and Future Work}
\label{sec:discussion_limitations}

First, this study utilizes dual-polarimetric $H/\alpha/A$ features derived from the $2\times 2$ covariance matrix $\mathbf{C}_2$ of Sentinel-1 observations. Future work will compare dual-polarimetric eigen-decomposition with quad-polarimetric $\mathbf{T}_3$ decomposition and model-based decompositions (Freeman--Durden and Yamaguchi) within self-supervised pre-training frameworks.

Second, our current experiments train and evaluate models within the acquisition periods of each respective study region. Future research will investigate cross-year and cross-regional transfer settings to examine how inter-annual shifts in planting dates, precipitation, and field management affect pre-trained representations, alongside rapid adaptation protocols using minimal target-year labels.

Third, label noise and fixed thresholding remain important practical considerations. Pixel-level CDL errors and boundary mixing affect USA evaluations, while boundary mismatches and management variations can still influence parcel-level datasets. Moreover, our Estonia parcel mapping applies a fixed majority-voting threshold ($r_p=0.6$); future work will integrate adaptive probability calibration and predictive uncertainty estimation for dynamic quality control.

Finally, while this study focuses on unimodal SAR time series, future work will incorporate optical canopy reflectance to construct multimodal SAR--optical pre-training frameworks for complex cropping systems.

\section{Conclusion}
\label{sec:conclusion}

To mitigate label dependency and fully exploit radar scattering physics in SAR time-series crop classification, we proposed SAR-TS-BERT, a physics-guided self-supervised pre-training framework. SAR-TS-BERT jointly models backscatter intensity features (VV, VH, and VV/VH) and $\mathbf{C}_2$-derived dual-polarimetric Cloude--Pottier decomposition parameters ($H/\alpha/A$) through a dual-branch cross-modal encoder with bidirectional cross-attention, and introduces a physics-guided contiguous span masking strategy (SpanPhysics) driven by temporal backscatter change rates to embed both polarimetric scattering mechanisms and phenological transition dynamics into self-supervised representation learning. Across independent test sets in Estonia (EuroCropsML), the USA (California CDL), and Chongming Island (Shanghai), SAR-TS-BERT achieves state-of-the-art overall accuracies of 96.78\%, 87.29\%, and a mean monthly OA of 99.37\%, respectively. Controlled ablation studies and Cohen's $d$ separability analyses confirm that backscatter and polarimetric decomposition features provide temporally asynchronous, complementary discriminative cues, and that grouped dual-branch encoding, bidirectional cross-attention, and SpanPhysics masking each contribute consistent incremental gains.

The benefits of physics-guided pre-training are particularly pronounced under severe label scarcity, during rapid phenological turnover, and in operational spatial mapping—improving OA and Macro F1 by 2.58 and 5.19 percentage points using only 1\% labeled data on Estonia, boosting Macro F1 by 1.45 and 1.44 percentage points during the March--April crop rotation transition window on Chongming Island, and enhancing intra-parcel agreement and wall-to-wall raster continuity. Collectively, these results establish that integrating SAR polarimetric scattering physics with temporal self-supervised pre-training effectively reduces annotation requirements and improves crop mapping reliability across diverse agricultural landscapes. Future work will extend SAR-TS-BERT to cross-year and cross-regional transfer scenarios and further investigate cross-branch attention interactions across phenological stages.

\bibliographystyle{IEEEtran}
\bibliography{references}

\end{document}